\documentclass{tlp}

\usepackage{xspace}

\usepackage{url}
\usepackage{graphicx}
\usepackage{afterpage}
\usepackage{amsmath}
\usepackage{subfigure}
\usepackage{lscape}
\usepackage{paralist} 

\usepackage{rotating}
\usepackage{bigstrut}
\usepackage{multirow}
\usepackage{booktabs}
\usepackage{float}

\usepackage{framed}
\usepackage{amssymb}
\usepackage{latexsym}

\def\pPointsT{100\xspace}

\newcommand\nameFont[1]{{\fontencoding{OT1}\fontfamily{cmss}\selectfont{#1}}}
\def\nf{\nameFont}

\def\aspcore{\nf{ASP-Core}\xspace}

\def\team{M\&S\xspace}

\def\solver{System\xspace}
\def\system{\solver}

\def\systemS{\ensuremath{\cal S}\xspace}
\def\trackT{\ensuremath{\cal T}\xspace}

\def\gbie{Grammar-based Information Extraction}

\newcommand\quo[1]{{``#1"}}
\newcommand\nop[1]{}
\newcommand\naf{{\ensuremath{{\mathbf{ not\ }}}}}

\newcommand{\p}{\ensuremath{{P}}}

\newcommand{\GP}{\ensuremath{grnd}(\p)\xspace}
\newcommand{\BP}{\ensuremath{B_{\p}}\xspace}
\newcommand{\UP}{\ensuremath{U_{\p}}\xspace}

\newenvironment{indentnew}[1]%
{\begin{list}{}%
         {\setlength{\leftmargin}{#1}}%
         \item[]%
} {\end{list}}

\def\clasp{{\tt \small clasp}}
\def\claspD{{\tt \small claspD}}
\def\Clingcon{{\tt \small Clingcon}}
\def\Clingo{{\tt \small Clingo}}
\def\iClingo{{\tt \small iClingo}}
\def\Gringo{{\tt \small Gringo}}
\def\Aclasp{{\tt \small Aclasp}}
\def\IDP{{\sc idp}}
\def\claspfolio{{\tt \small claspfolio}}
\def\cmodels{{\sc cmodels}}
\def\fastdownward{{\em fastdownward}}

\newcommand\mycite[1]{\cite{#1}} 

\begin{document}

\submitted{24 August 2011}
\revised{12 January 2012}
\accepted{4 June 2012}

\title[The Third Open Answer Set Programming Competition]
{The Third Open Answer Set Programming Competition}

\author[F. Calimeri, G. Ianni and F. Ricca]
{
FRANCESCO CALIMERI, GIOVAMBATTISTA IANNI, FRANCESCO RICCA \\
Dipartimento di Matematica \\
Universit\`{a} della Calabria, Italy
\email{\{calimeri,ianni,ricca\}@mat.unical.it}
}

\setcounter{page}{1}

\maketitle

\label{firstpage}

\begin{abstract}
\nopagebreak
Answer Set Programming (ASP) is a well-established paradigm of declarative
programming in close relationship with other declarative formalisms
such as SAT Modulo Theories, Constraint Handling Rules, FO(.),
PDDL and many others.
Since its first informal editions, ASP systems
have been compared in the now well-established ASP Competition. The Third (Open) ASP
Competition, as the sequel to the ASP Competitions Series held at the
University of Potsdam in Germany (2006-2007) and at the University of Leuven
in Belgium in 2009, took place at the University of Calabria (Italy) in the
first half of 2011.
Participants competed on a pre-selected collection of benchmark problems, taken from a variety of domains as
well as real world applications.

The Competition ran on two tracks: the Model and Solve (\team) Track, based on an open problem encoding,
and open language, and open to any kind of system based on a declarative specification paradigm; and
the \solver Track, run on the basis of fixed, public problem encodings, written in a standard ASP language.
This paper discusses the format of the Competition and the rationale behind it, then
reports the results for both tracks. Comparison with the second ASP competition and
state-of-the-art solutions for some of the benchmark domains is eventually discussed.
\end{abstract}

\section{Introduction}\label{sec:introduction}
\nopagebreak
Answer Set Programming (ASP) is a declarative approach to computer programming stemming roots in the area
of nonmonotonic reasoning and logic programming
\mycite{gelf-lifs-91,niem-99,mare-trus-99}.
The main advantage of ASP\footnote{For introductory material on ASP, the reader might refer to
\mycite{bara-2002,eite-etal-2009-primer}.} is its high declarative nature combined with a
relatively high expressive power~\mycite{dant-etal-01}.
After some pioneering work
\mycite{bell-etal-94,subr-etal-95}, nowadays there are a number of
systems that support ASP and its variants
\mycite{ange-etal-2005-lpar,dalp-etal-2009-fi,gebs-etal-2007-ijcai,janh-niem-2004,lefe-nico-2009-lpnmr,leon-etal-2002-dlv,lier-mara-2004-lpnmr,lin-zhao-2004,simo-etal-2002}.
The availability of some efficient systems make ASP a powerful tool
for developing advanced applications in several fields, ranging from
Artificial Intelligence~\mycite{bald-etal-01,bara-gelf-2000,bara-uyan-2001,frie-08-techrep,fran-etal-2001,noge-etal-2001,wasp-showcase}
to Information Integration~\mycite{leon-etal-2005,caniup-bertossi-2010},
Knowledge Management~\mycite{bara-2002,bard-95,grass-etal-09-apps-lpnmr},
Bioinformatics~\mycite{palo-etal-2005-ISMIS,dovier-2011,gebs-etal-2011-tplp-bio}, and
has stimulated some interest also in industry~\mycite{grass-etal-2010-a,ricca-etal-2010-idum}.

ASP systems are evaluated in the now well-established ASP Competitions, that
started with two informal trials at ASP Dagstuhl meetings in 2002 and 2005.
The present competition, held at the University of Calabria (Italy), is the third official edition,
since the rules of the contest were formalized and implemented in the first two ``official'' ASP
Competitions \mycite{gebs-etal-2007-lpnmr-competition,2nd-ASP-competition}.
Besides comparing ASP systems with each other, one of the goals of the competition is to benchmark similar systems and declarative paradigms close in spirit to ASP.
To this end, the Third ASP Competition featured two tracks: the Model and Solve Competition Track (\team Track from now on), based on an open problem encoding,
open language basis, and open to any system based on a declarative specification paradigm; and
the \solver Competition Track, based on a fixed problem encodings, written in a standard ASP language.
The \team Competition Track  essentially follows the direction of the previous ASP Competition; the
\solver Competition Track (\solver Track from now on) was conceived in order to compare participant systems on the basis
of fixed input language and fixed conditions.

A preliminary work reporting results of the \solver Track appeared in \citeNP{lpnmr11a};
this paper extends that work in the following respects:

\begin{itemize}
\item
detailed results of the \solver Track, which now include non-participant systems such
as parallel solvers and some latecomers.
\item
description of the problem categories, including those appearing in the \team Track only;
\item
discussion of the rationale and the rules of the \team Track, presentation of competitors and results of the Track;
\item
a number of comparisons; namely, we show:
\begin{itemize}
\item how the winner of the Second ASP Competition performed on this edition's benchmarks;
\item whenever applicable, how participants to this edition performed on the former Competition benchmarks;
\item for systems to which this is applicable, whether and how performance changed when switching from the \solver Track settings
to the more liberal settings of the \team competition;
\item
for a selection of benchmarks, how the participants performed against some known state-of-the-art solutions;
these ranged from specialized algorithms/systems to tailored ad-hoc solutions based on constraint programming and/or SAT.
\end{itemize}
\end{itemize}

The remainder of the paper is structured as follows: in Section~\ref{sec:competition-format}
we discuss the Competition rationale, the subsequent format and regulations for both Tracks, and
we briefly overview the standard language adopted in the \solver Track;
Section~\ref{sec:participants} illustrates the classes of declarative languages participating in the \team Track
and the classes of evaluation techniques adopted (particularly focusing on the \system Track), and presents the participants in the Competition;
in Section~\ref{sec:compsettings} we illustrate the scoring criteria, the benchmark suite and other competition settings;
Section~\ref{sec:results} reports and discusses the actual results
of the Competition; in Section \ref{sec:participantsToComparisonsWithYardsticks} we report details about comparisons of participants with a number of yardstick problem solutions and/or systems;
conclusions are eventually drawn in Section \ref{sec:conclusions}. An electronic appendix details the above whenever appropriate.

\section{Competition Format}\label{sec:competition-format}
\nopagebreak
In this Section we describe the Competition format for the \solver and \team Track,
thoroughly discussing motivations and purposes that led to the choices made:
the two tracks differ in regulations and design principles. It must be observed that the \solver Track resembles competitions of neighboring communities in spirit, and is indeed played on a fixed input language, with fixed input problem specifications. However, both tracks introduce specific aspects related to the ASP philosophy. As a main difference,
note that competitions close to the ASP community (e.g. SAT, CASC, IPC) are run on a set of couples $(i,S)$, for $i$ an input instance and $S$ a participant solver. Instead, the ASP competition has problem specifications as a further variable. ASP Competitions can be seen as played on a set of triples $(i,p,S)$: here, $i$ is an input instance, $p$ a problem specification, and $S$ a solver. Depending on track regulations, $i$, $p$ and $S$ are subject to specific constraints.

\paragraph{\bf \solver Track Format.}\label{subsec:system-format}

The regulations of the \solver\ Track were conceived taking into account two main guidelines.
As a first remark, it must be observed that ASP is still missing a {\em standard} high-level input language, in contrast with
other similar declarative paradigms.\footnote{These range from the Satisfiability Modulo Theories SMT-LIB format \mycite{smt-lib-web}, the Planning Domain Definition Language (PDDL) \mycite{pddl-techrep-2005}, the TPTP format used in the CASC Automated Theorem Proving System Competitions \mycite{cade-atp-web}, to the Constraint Handling Rules (CHR) family \mycite{chr-web}.} It was thus important to play on the grounds of a common language, despite restriction to commonly acknowledged constructs only.
As a second guideline,
it has been taken into account that the outcome of the \solver Track should give a
fairly objective measure of what one can expect when switching from one system to another,
while keeping all other conditions fixed, such as the problem encoding and solver settings.
In accordance with the above, the \solver Track was held on the basis of the following rules.
\begin{enumerate}
\item
    The Track was open to systems able to parse input written
    in a fixed language format, called \aspcore;
\item
    For each benchmark problem, the organizers chose a fixed \aspcore specification:
    each system had to use this specification compulsorily for solving the problem at hand;
\item
    Syntactic special-purpose solving techniques, e.g. recognizing a problem from file names, predicates name
    etc., were forbidden.
\end{enumerate}

The detailed rules and the definition of \quo{syntactic technique} are reported in \ref{app:competitionsettings}.

\paragraph{The language \aspcore.}
\aspcore is a rule-based language its syntax stemming from plain Datalog and Prolog:
it is a conservative extension to the nonground case of the {\em Core} language adopted in the First ASP Competition;
it complies with the core language draft specified at LPNMR 2004 \mycite{draft2004},
and refers to the language specified in the seminal paper \mycite{gelf-lifs-91}.
Its reduced set of constructs is nowadays common for ASP parsers and can be supported
by any existing system with very minor implementation effort.%
\footnote{During competition activities, we also developed a larger language proposal, called ASP-RfC (Request for Comments),
including aggregates and other widely used, but not yet standardized, language features.}
\aspcore features disjunction in the rule heads, %
both strong and negation-as-failure (NAF) negation in rule bodies, as well as nonground rules.
A detailed overview of \aspcore is reported in \ref{app:aspcoreformat};
the full \aspcore language specification can be found in \mycite{languageformat}.

\paragraph{\bf Model and Solve Track Format.}\label{subsec:m&sCompetition-format}

The regulations of the \team Track take into account
the experience coming from the previous ASP Competitions.
As driving principles in the design of the \team Track regulations we can list:
encouraging the development of new expressive declarative constructs and/or new modeling paradigms;
fostering the exchange of ideas between communities in close relationships with ASP; and, %
stimulating the development of new ad-hoc solving methods, and refined problem
specifications and heuristics, on a per benchmark domain basis.
In the light of the above, the \team Track was held under the following rules:
\begin{enumerate}
\item
    the competition organizers made a set
    of problem specifications public, together with a set
    of test instances, these latter expressed in a common instance input format;
\item
    for each problem, teams were allowed to submit a
    specific solution bundle, based on a solver (or a combination of solvers) of choice, and a
    problem encoding;
\item
    any submitted solution bundle was required
    to be mainly based on a declarative specification language.
\end{enumerate}

\section{Participants}\label{sec:participants}
\nopagebreak
In this section we present the participants in the competition, categorized by
the adopted modeling paradigms and their evaluation techniques.

\subsection{\solver Track}\label{subsec:participantsToSystem}
\nopagebreak
The participants in the \solver Track were only ASP-Based systems.
The traditional approach to ASP program evaluation follows an instance processing work-flow
composed of a grounding module, generating a propositional theory, coupled with a subsequent propositional solver module.
There have been other attempts deviating from
this customary approach \mycite{dalp-etal-2009-fi,lefe-nico-2009-lpnmr-2,lefe-nico-2009-lpnmr};
nonetheless all the participants adopted the canonical \quo{ground \& solve} strategy.
In order to deal with nonvariable-free programs, all solvers eventually relied on the grounder \Gringo~\mycite{gebs-etal-2007-gringo}.
In detail, the \system Track had eleven official participants and five noncompeting systems.
These can be classified according to the employed evaluation strategy as follows:

\paragraph{Native ASP:} featuring custom propositional search techniques
that are based on backtracking algorithms tailored for dealing with logic programs.
To this class belong:  \clasp~\mycite{gekasc09b},
\claspD~\mycite{dres-etal-2008-KR}, \claspfolio~\mycite{gebs-etal-2011-claspfolio},
\Aclasp~\mycite{aclasp-web}, {\sc Smodels}~\mycite{simo-etal-2002}, \IDP~\mycite{witt-etal-2008-idp},
and the non-competing \clasp-{\tt \small mt}~\mycite{ellg-etal-2009-clasp-mt}.
\clasp\ features techniques from the area of boolean constraint solving,
and its primary algorithm relies on conflict-driven nogood learning.
\claspD\ is an extension of \clasp\ that is able to solve unrestricted disjunctive logic programs,
while \claspfolio\ exploits machine-learning techniques in order to choose
the best-suited configuration of \clasp\  to process the given input program;
\Aclasp\ (non-participant system) is a variant of clasp employing a different restart strategy;
\clasp-{\tt \small mt} is a multi-threaded version of \clasp.
\IDP\ is a finite model generator for extended first-order logic theories.
Finally, {\sc Smodels}, one of the first robust ASP systems that have been made available to the community,
was included in the competition for comparison purposes, given its historical importance.

\paragraph{SAT-Based:} employing translation techniques --e.g., completion \mycite{fage-94},
loop formulas \mycite{lee-lifs-2003,lin-zhao-2004}, nonclausal constraints \mycite{lier-2008-sup}--
to enforce correspondence between answer sets and satisfying assignments of SAT formulas
so that state-of-the-art SAT solvers can be used for computing answer sets.
To this class belong: \cmodels~\mycite{lier-mara-2004-lpnmr},  {\sc sup}~\mycite{lier-2008-sup}
and three variants of {\sc lp2sat}~\mycite{janh-2006-journal-lp2sat}: ({\sc lp2gminisat}, {\sc lp2lminisat} and {\sc lp2minisat}).
In detail, \cmodels\ can handle disjunctive logic programs and
exploits a SAT solver as a search engine for enumerating models, and also
verifying model minimality whenever needed; {\sc sup}\ makes use of nonclausal constraints, and
can be seen as a combination of the computational ideas behind \cmodels\ and {\sc Smodels};
the {\sc lp2sat} family of solvers, where the trailing $g$ and $l$ account
for the presence of variants of the basic strategy, employed {\sc MiniSat}~\mycite{nikl-nikl-2003-minisat}.

\paragraph{Difference Logic-based:} exploiting a translation \mycite{janh-etal-2009-lp2diff}
from ASP propositional programs to Difference Logic (DL) theories \mycite{nieu-oliv-2005-difference}
to perform the computation of answer sets via Satisfiability Modulo Theories \mycite{nieu-etal-2006-sat} solvers.
To this class belongs {\sc lp2diffz3} and its three non-competing variants, namely: {\sc lp2diffgz3}, {\sc lp2difflz3}, and {\sc lp2difflgz3}.
The {\sc lp2diff} solver family~\mycite{janh-etal-2009-lp2diff} translates ground ASP programs
into the QF$\!\_\,$IDL dialect (difference logic over integers) of the SMT library \mycite{smt-lib-web};
the trailing $g$, $l$ and $lg$ letters account for different variants of the basic translation technique.
The {\sc lp2diff} family had {\em Z3}~\mycite{mend-bjor-2008-z3}, as underlying SMT solver.

\subsection{\team Track}\label{sec:teampart}
\nopagebreak
The \team Competition Track was held on an open problem encoding, open language basis.
Thus participants adopted several different declarative paradigms,
which roughly belong to the following families of languages:
{\em ASP-based}, adopting ASP~\mycite{gelf-lifs-91} (and variants) as modeling language;
{\em FO(.)-based}, employing FO(ID)~\mycite{dene-tern-2008-TOCL};
{\em CLP-based}, using logic programming as declarative middle-ware language for reasoning on constraint satisfaction problems
\mycite{jaffar1987}; and, {\em Planning-based}, adopting PDDL (Planning Domain Definition Language) as modeling language \mycite{PDDL-changes-3.1-web-2008}, respectively.
In detail, six teams participated to the \team Track:

\paragraph{\bf Potassco:}
The Potassco team from the University of Potsdam, Germany \mycite{potassco-web}
submitted a heterogenous ASP-based solution bundle.
Depending on the benchmark problem, Potassco employed
\Gringo~\mycite{gebs-etal-2007-gringo} coupled with either \clasp~\mycite{gekasc09b} or \claspD~\mycite{dres-etal-2008-KR},
and \Clingcon, which is an answer set solver for constraint logic programs, built upon the \Clingo\ system and the CSP solver Gecode \mycite{gecode-web}, embedding and extending \Gringo\  for grounding.

\paragraph{\bf \Aclasp:}
The team exploited the same ASP-based solutions provided by the Potassco team,
and participated only in a number of selected problem domains.
The solver of choice was \Aclasp~\mycite{aclasp-web}, a modified version of \clasp\
that features a different restart-strategy, depending on the average decision-level
on which conflicts occurred. The grounder of choice was \Gringo.

\paragraph{\bf \IDP:}
The \IDP~\mycite{witt-etal-2008-idp} team, from the Knowledge Representation and Reasoning (KRR)
research group of K.U.Leuven, Belgium, proposed FO(.)-based solutions.
In particular, problem solutions were formulated in the $FO(.)$ input language,
and the problem instances were solved by {\sc MiniSatID}\ 
\mycite{maar-etal-2008-minisatid}
on top of the the grounder {\sc Gidl}\ 
\mycite{witt-etal-2008-gidl}.
A preprocessing script was used to rewrite ASP instances into FO(.) structures.

\paragraph{\bf EZCSP:}
EZCSP is an Eastman Kodak Company and University of Kentucky joint team.
The team is interested in evaluating and comparing ASP and hybrid languages
on challenging industrial-sized domains.
EZCSP \mycite{balduccini2009representing} is also the name of both the
CLP-based modeling language, featuring a lightweight integration between ASP
and Constraint Programming (CP), and the solver employed by this team.
The EZCSP system supports the free combination of different ASP and CP solvers
which can be selected as sub-solvers, according to the features of the target domain.
In particular, the team exploited
the following ASP solvers depending on the benchmark at hand:
\clasp, \iClingo\ and ASPM \mycite{balduccini2009AGeneral}.
Moreover, in cases where CP constraints were used,
the team selected B-Prolog as solver.%

\paragraph{\bf BPSolver:}
This team adopted a CLP-based modeling language and exploited the B-Prolog system \mycite{Zhou-2011-tplp}
for implementing solutions.
BPSolver employed either pure Prolog, tabling techniques \mycite{chen-warr-1996-tabling},
or CLP(FD) \mycite{vanh-2989-dlp-fd} depending on the problem at hand.
In particular, apart from a few problems that required only plain Prolog, all
the provided solutions were based on either CLP(FD) or tabling.

\paragraph{\bf Fast Downward:}
This is an international multiple research institutions joint team
that proposed some planning-based solutions.
Benchmark domains were statically modeled as planning domains,
and problem instances were automatically translated from ASP to PDDL.
The submitted solutions were based on Fast Downward \mycite{Helm-2006-fast-downward},
a planning system developed in the automated planning community.
The team restricted its participation to benchmarks that could be easily seen as planning problems
and exploited a number of different configurations/heuristics of Fast Downward.

\section{Competition Settings}\label{sec:compsettings}
\nopagebreak
We now briefly describe the scoring methodology of choice, the selected benchmarks and other practical settings in what
the competition was run. A detailed description of general settings, the scoring criteria and the selected
benchmark suite can be respectively found in Appendices B, C and D.

\paragraph{Scoring System.}
The scoring framework is a refinement of the one adopted in the first and second ASP Competitions. In these former editions, scoring rules were mainly based on a weighted sum of the number of instances solved within a given time-bound; in this edition, the scoring framework has been extended by awarding additional points to systems performing well in terms of evaluation time.
For search and query problems, each system on benchmark problem $P$ was awarded the score
$S(P) = S_{solve}(P)+ S_{time}(P)$.
$S_{solve}$ and $S_{time}$ could range from $0$ to $50$ each:
while $S_{solve}$ is linearly dependent on the number of instances solved in the allotted time,
$S_{time}$ contains a logarithmic dependence on participants' running times, thus making
less significant time differences in the same order of magnitude.
As for optimization problems, the $S_{solve}$ quota was replaced with a scoring formula taking into account also the solution quality,
in particular, the closer to the optimal cost, the better exponentially.

\paragraph{Benchmark suite.}
There were a total of 35 selected benchmark domains,
mainly classified according to the computational complexity of the related problem,
in {\em Polynomial}, {\em NP}, and {\em Beyond-NP} ones, where this latter category
was split into {\em $\Sigma^P_2$} and {\em Optimization}. The benchmark suite included
planning domains, temporal and spatial scheduling problems, combinatory puzzles,
a few graph problems, and a number of applicative domains taken from the database,
information extraction and molecular biology field. According to their type, problems were
also classified into {\em Search}, {\em Query} and {\em Optimization} ones.

\paragraph{Software and Hardware Settings.}
The Competition took place on servers featuring a 4-core Intel Xeon CPU X3430 running at 2.4 Ghz, with 4 GiB of physical RAM.
All the systems where benchmarked with just one out of four processors enabled, with the exception
of the parallel solver \clasp-{\tt \small mt},
and were allowed to use up to 3 GiB of user memory. The allowed execution time for each problem's instance was set at
$600$ seconds. 

\section{Results and Discussion}\label{sec:results}
\nopagebreak
The final competition results are reported in Figures \ref{fig:table-system-categories} and \ref{fig:table-team-categories}, for \system
and \team Track, respectively.
The detailed results for each considered benchmark problem, and
cactus plots detailing the number of instances solved and the corresponding
time, on a per participant basis, are reported in \ref{app:results}
(note that timed out instances are not drawn).
Full competition figures, detailed on a per instance basis, together with executable packages and declarative
specifications submitted by participants, are available on the competition web site \cite{aspcomp2011-web}.

\subsection{\solver Track Results}\label{subsec:resultsSystem}
\nopagebreak
\paragraph{Polynomial Problems.}\label{subsec:presultsSystem}
\nopagebreak
Grounding modules are mainly assessed while dealing with problems from this category, with the notable exception
of two problems, for which, although known to be solvable in polynomial time,
we chose their natural declarative encoding, making use of disjunction.
In the case of these last two problems,
the \quo{combined} ability of grounder and propositional solver modules was tested.
The aim was to measure whether, and to what extent, a participant system
could be able to converge on a polynomial evaluation strategy when
fed with such a natural encoding. All
the participant systems employed \Gringo\ (v.3.0.3) as the grounding module: however, we noticed some systematic performance differences, owing to the different command line options fed to \Gringo\  by participants. The winner of the category is \clasp, with $213$ points,
as shown in Figure \ref{fig:table-system-categories}.
Interestingly, Figure \ref{fig:plots-detail-p} of \ref{app:results} shows a sharp difference between a group of easy and hard instances: notably, these latter enforced a bigger memory footprint when evaluated; indeed, it is worth mentioning that the main cause of failure in this category was out of memory, rather than time-out. Instances were in fact relatively large, usually.

\paragraph{NP Problems.}\label{subsec:npresultsSystem}
The results of this category show how \claspfolio\ ($609$ points) slightly outperformed \clasp\ and \IDP\ ($597$ points), these
latter having a slightly better time score ($227$ versus $224$ of \claspfolio).

\paragraph{Beyond-NP Problems.}\label{subsec:bnpresultsSystem}
Only the two systems \claspD\ and \cmodels\ were able to deal with the two problems in this category, with \claspD\ solving
and gaining points on both problems, and \cmodels\ behaving well on {\sc MinimalDiagnosis} only.

\paragraph{Overall Results.}\label{sec:totalresultsSystem}
Figure \ref{fig:table-system-categories} shows \claspD\   as the overall winner, with $861$ points:
$560$ points were awarded for the instance score, corresponding to a total of $112$ instances
solved out of $200$. \claspfolio\ and \clasp\   follow with a respective grandtotal of $818$ and $810$.
It is worth noting that \claspD\   is the only system, together with \cmodels, capable
of dealing with the two Beyond-NP problems included in the benchmark suite, this giving to \claspD\
a clear advantage in terms of score.

\paragraph{Non-competing Systems.}\label{subsec:noncomp}
After the official competition run, we additionally ran five non-competing systems,
including: the parallel system \clasp-{\tt \small mt}, and a number of solvers which
did not meet the final official deadline.
We included here also those systems in order to give a wider
picture of the state of the art in ASP solving.

Noncompeting systems are reported in italics in
Figure \ref{fig:table-system-categories},
and their behavior is plotted in
Figures 5-8 of \ref{app:results},
with the winner of the \solver Track used as a yardstick.
Note that noncompeting executables were mostly variants of
the executables presented by the Potassco and the Aalto teams.
None of them performed clearly better than the ``official'' participating versions.
The best sequential noncompeting system was \Aclasp, with a score of 780 points,
which would have reached the fifth absolute position in the final classification,
see Figure \ref{fig:table-system-categories}.

A special mention goes to the parallel system \clasp-{\tt \small mt},
the only system that ran on a machine with four CPUs enabled.
\clasp-{\tt \small mt} is a comparatively young system and (although it was disqualified from some
domains\footnote{The {\em overall score} for a problem $P$ is set to zero if the system produces an incorrect
answer for some instance $P$. See \ref{app:detailedscoring} for more details.})
it is clearly the best performer in $NP$ totalizing 629 points in this category
corresponding to 29 points more than the best sequential system (\claspfolio).

This result confirms the importance of investing in parallel solving techniques
for exploiting the nowadays-diffused parallel hardware.

\subsection{\team Track Results}\label{subsec:resultsTeam} 
\begin{figure}[t!] 
\begin{minipage}[b]{\textwidth} \centering
  \includegraphics[width=.899\textwidth]{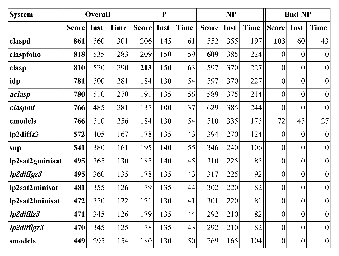}
\vspace{-0.2cm}
 \caption{\system Track - Results by Categories}\label{fig:table-system-categories}
\vspace{0.4cm}
\end{minipage}
\begin{minipage}[b]{\textwidth} \centering
 \centering
  \includegraphics[width=.899\textwidth]{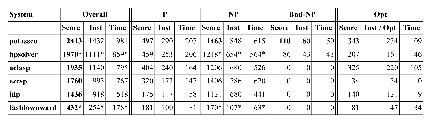}
\vspace{-0.2cm}
 \caption{\team Track - Results by Categories}\label{fig:table-team-categories}
\end{minipage}
\end{figure}

\vspace{-0.1cm}
\nopagebreak
%
\paragraph{Polynomial Problems.}\label{subsec:presultsTeam}
The winner in the category (see Figure~\ref{fig:table-team-categories}) is the Potassco team. It is worth mentioning that the runner-up BPsolver,
which for this category presented solutions based on predicate tabling, was the absolute winner in three out of
the seven problem domains. This suggests that top-down evaluation techniques might pay off, especially for polynomial
problems.

\paragraph{NP Problems.}\label{subsec:npresultsTeam}
The category shows the Potassco team as winner with 1463 points, closely followed by EZCSP (1406 points), which was
notably the best performing team on {\sc ReverseFolding} and {\sc Packing}. Also, BPsolver was by far the fastest system in six domains out of nineteen, although its
performance was fluctuating (e.g. in {\sc Solitaire} and {\sc GraphColouring}) and it was disqualified in a couple
of domains.\footnote{For the sake of scientific comparison, Figure \ref{fig:table-team-totals} reports also scores
obtained by bpsolver on a late submission for {\sc HanoiTowers}, fixing the faulty solution submitted within the deadline. Grand totals include this latter score.}
The very good performance of \IDP\ on {\sc GraphColouring} is also worth mentioning.

\paragraph{Beyond-NP Problems.}\label{subsec:bnpresults}
Only two teams submitted solutions for the two problems in this category, with the Potassco team being the clear winner in both domains.

\paragraph{Optimization Problems.}\label{subsec:optresults}
In this category the two clasp-based solution bundles (Potassco and \Aclasp) outperformed the rest of participants, with \IDP\ being the first nonclasp-based system in the category. As in the NP category, BPsolver was the best team in a couple of domains.

\paragraph{Overall Results.}\label{sec:totalresultsTeam}
Figure~\ref{fig:table-team-categories} shows the Potassco solution bundle as the clear
winner of the \team Track (see Figure~\ref{fig:plots-overall-team}).
The results detailed per benchmark (see Appendix F, Figure~\ref{fig:table-team-totals})
show that all the teams were best performers in one ore more domains with very encouraging results.

A special remark must be made concerning the \fastdownward\ team, coming
from the planning community. It competed in only
a few number of domains, mostly corresponding to planning problems,
totalizing 433 points. Given the small number of problems which \fastdownward\
participated in, a clear comparison cannot be drawn: however, it can be
noted that \fastdownward\ performed quite well on {\sc HydraulicLeaking} and {\sc HydraulicPlanning}.
In other domains, the relatively low performance can be explained considering
that some of the problems where specified more in the form of knowledge
representation problems, with some common restrictions specific to ASP.

The solutions proposed by all participants were very heterogenous, ranging
from purely declarative to the usage of Prolog in a nearly procedural style.
Among the lessons learned, it is worth observing the fact that purely declarative
solutions very often paid off in terms of efficiency, and outperformed
comparatively more tweaked approaches to problem solving.

\section{Further Analysis}\label{sec:participantsToComparisonsWithYardsticks}
\begin{figure}[t]
 \centering
  \includegraphics[width=\textwidth]{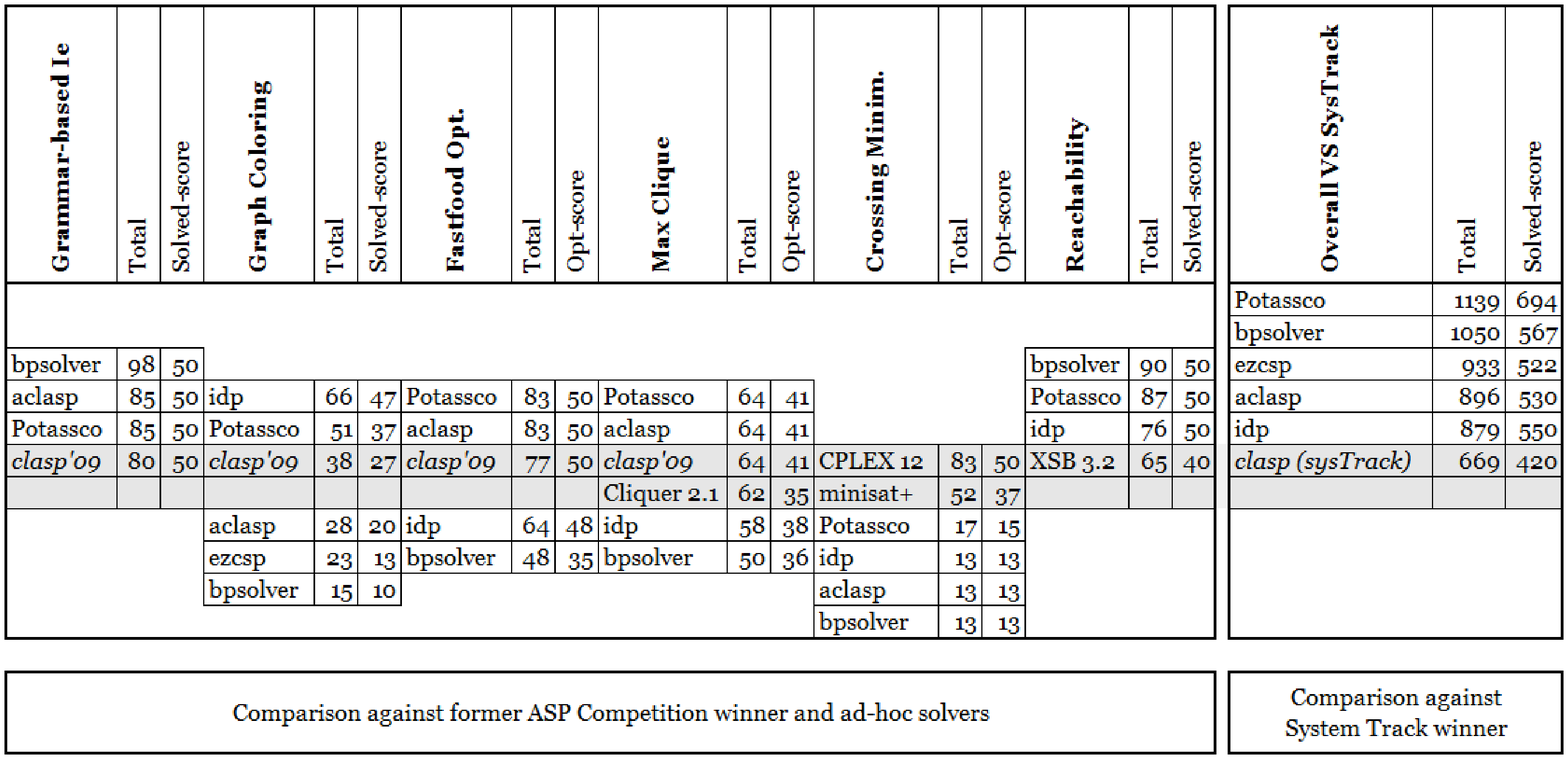}
 \caption{\label{fig:comparisons} Comparisons with State-of-the-art Solutions and other Yardsticks}
\end{figure}

\nopagebreak
In this section a number of additional analyses are reported, with the aim of
giving a clearer and more complete picture of the state of the art in ASP solving
and declarative programming.
In particular: we compared the winning solution bundles\footnote{As \quo{solution bundle} we mean
here the combination of ad-hoc tuned solver binaries together with ad-hoc encodings, as they were submitted in the second ASP competion.} submitted to the former ASP Competition
with the updated ones submitted also to the Third ASP Competition,
so that two years of advances in the state of the art are outlined;
we measured the distance in performance among ASP-based solutions and some specialized ad-hoc solutions
available in the literature, for some specific benchmark problems considered in the current competition;
and, eventually, we assessed the impact of fine-tuning of systems/solutions
by comparing specialized executables and problem encodings submitted to the \team Track with
a good-performing default-setting ASP system of the \solver Track.
The outcomes are summarized in Figure~\ref{fig:comparisons}.
A grey strip highlights the lines corresponding to \quo{yardstick} systems
which competitors have been compared to. {\em Total} scores are computed according
to the rules of this Competition;
{\em Solved-score}, which roughly corresponds to the score computed according to past Competition rules,
is obtained by subtracting the score corresponding to the time quota from the Total score introduced in this competition
(see \ref{app:detailedscoring} for more insights).
The results are discussed in detail in the following.

\paragraph{\bf The State-of-the-art after Two-years of Improvements.}\label{subsec:comparisonFormerASP-SOTA}
In order to assess possible improvements over former participants
in the Second ASP Competition, we selected some significant problems
appearing both in the Third and Second edition of the Competition
with same specification. This set of problems counts a
polynomial problem ({\sc \gbie}), a NP problem ({\sc GraphColoring}),
and two optimization problems ({\sc FastFoodOptimization} and {\sc MaximalClique}).
On these benchmarks, we ran the solution bundles
submitted by the winners of the 2nd ASP Competition (the Potassco team)
alongside all the solution bundles of the current participants
to the \team Track. The instance families used for this test were
composed of both the instances used in the 2nd ASP Competition and in
the current edition.

The state of the art in the last two years has been clearly pushed forward,
as witnessed by the results reported in the four leftmost sections of Figure~\ref{fig:comparisons}
(corresponding to the above-mentioned problems).
Indeed, the new solution bundles based on \clasp\ (indicated by {\em Potassco} in Figure~\ref{fig:comparisons}) outperformed in all considered benchmarks
the ones (indicated by {\em clasp'09} in Figure~\ref{fig:comparisons})
submitted to the Second ASP competition.
Note also that, other current solution bundles (i.e., bpsolver, \Aclasp, \IDP)
were often able to outperform {\em clasp'09}, and are generally comparable
to {\em Potassco} even beating it on the {\sc \gbie} and {\sc GraphColoring} problems.

\paragraph{\bf Participants vs Ad-hoc Solutions.}\label{subsec:comparisonSOTA-adHOC}
Participants in the Competition were based on declarative formalisms,
and were essentially conceived as \quo{general-purpose} solvers.
For a selection of benchmark domains, we compared participants in the \team Track
with specialized ad-hoc solutions, not necessarily based on declarative specifications,
with the aim of figuring out what a user might expect to pay in order
to enjoy the flexibility of a
declarative system.

\paragraph{Maximal Clique.} {\sc MaximalClique} is a graph
problem with a longstanding history of research towards
efficient evaluation algorithms. It was one of the problems
investigated in the early Second DIMACS Implementation Challenge \mycite{john-tric-1993-dimacs},
and research continued on the topic later on \mycite{gibb-etal-1996-maxclique,bomz-etal-1999-maxclique,guti-2004-graph}.
In the Competition, the problem was specified as
finding the maximum cardinality clique on a given graph \mycite{aspcomp2011-web}
and most of the instances were taken from BHOSLIB \mycite{bhoslib-web}.

We compared participants in the \team Track with
Cliquer \mycite{cliquer-web}.
Cliquer is an up-to-date implementation of an exact branch-and-bound algorithm,
which is expected to perform well while dealing with several classes of graphs
including sparse, random graphs and graphs with certain
combinatorial properties \mycite{oste-2002-fast-clique}.

Cliquer is based on an exact algorithm: in this respect we found it
the more natural choice for comparison with participants in the ASP
Competition, all of which are based on exhaustive search algorithms.
A comparison with other existing approximate algorithms
\mycite{bopp-hall-1992-indip-set-approx,feig-2005-clique-approx}
would have been expectedly unbalanced in favor of these latter.
Our findings show that, in the setting of the competition, {\em Potassco}  and \Aclasp\
have comparable performance with Cliquer, while \IDP\ performed quite close to them.

\paragraph{Crossing Minimization in layered graphs.}
Minimizing crossings in layered graphs is an important problem
having relevant impact e.g., in the context of VLSI layout optimization.
The problem has been studied thoroughly, and valuable algorithms have
been proposed for solving it, among which \mycite{jung-etal-1997-crossingminimization,heal-kuus-1999-crossingminimization,mutz-2000-crossingminimization} and \cite{gang-etal-2010-crossingminimization}.
The instances considered for the Competition were taken from the graphviz repository
\mycite{graphviz-web}.

We ran the two ad-hoc solutions proposed in \mycite{gang-etal-2010-crossingminimization}
over the \team Track instance family and compared results
with outcomes of participants in the \team competition.
The two solutions were, respectively, based on a translation to SAT and to
Mixed Integer Programming. The former was run using {\sc MiniSat}+
\mycite{een-sore-2006-minisatplus} as solver, while the latter used
CPlex 12.0 \mycite{cplex-web}. Both solvers were run using default settings.

The comparison shows a big gap between participants in
the competition and the two yardsticks, which both perform much better.

\paragraph{Reachability.} This polynomial problem is a distinctive
representative of problems that can be naturally specified
using recursion in plain logic programming, lending itself to
comparison with other logic programming-based systems.

We compared the outcomes of participants in the \team Track
with XSB 3.2 \mycite{xsb-web},
one of the reference systems of the OpenRuleBench initiative. OpenRuleBench
\mycite{fodo-etal-2011-openrulebench-report} is aimed at
comparing rule-based systems both from the Deductive Database
area, Prolog-based and oriented to RDF triples.

All the three participants submitting a solution to the {\sc Reachability}
problem outperformed XSB, especially in terms of time performance. However,
it must be stated that the scoring system of the ASP Competition purposely does
not exclude loading and indexing steps when measuring execution times,
differently from the setting of OpenRuleBench; furthermore, XSB
was run with its {\em off-the-shelf} configuration, except for
attribute indexing and tabling appropriately enabled. Indeed, BPsolver
depends here on the same technology as XSB (tabling and top down),
but fine-tuned to the problem.

\paragraph{\bf The Effects of Fine-tuning.}\label{subsec:comparisonWOSystem}
Recall that the setting of the \system Track prevented all participants
from developing domain-dependent solutions and,
on the contrary, the \team Track allowed the submission
of fine-tuned encodings and the static selection of systems parameters/heuristics.

In order to assess the impact of fine-tuning,
we selected the \clasp\ version as yardstick that participated
in the \system Track, labelled {\em clasp (sysTrack)} in Figure~\ref{fig:comparisons}; then we
ran it over $P$ and $NP$ problems which were in common between the \system Track and the \team Track problem suites.%
\footnote{The {\sc HydraulicPlanning} benchmark has been excluded since its specification was
different in the \team Track.}
{\em clasp (sysTrack)} was run using the fixed \aspcore encodings and the settings of the \system Track, but over the different (and larger) instance sets coming from the \team Track.
These choices ensured a comparison more targeted to the assessment of the impact of tuning: indeed,
$(i)$ the {\em clasp (sysTrack)} executable is the ``naked''%
\footnote{In the sense that it does not feature any parameter tuning technique.}
solver of choice in almost all the solution bundles submitted by the Potassco team, the winner of the \team Track;
$(ii)$ {\em clasp (sysTrack)} is the winner of the \system Track in the $P$ category, and runner-up in the $NP$ category (in both it performed better than the overall winner \claspD);
$(iii)$ there is no significant difference between the solutions presented by the Potassco team in both the \system and the \team Track for $Beyond NP$ problems.

The obtained results are reported in Figure~\ref{fig:comparisons}, rightmost side,
and, as expected, confirm the importance of fine-tuning and customized encodings.
Indeed, \quo{tuned} solution bundles outperformed
the fixed configuration of {\em clasp (sysTrack)}.
This clearly indicates the need for further developing new optimization techniques
and self-tuning methods (e.g., on the line of \mycite{gebs-etal-2011-claspfolio}),
and to further extend the basic standard language, in order to
make efficient ASP solutions within reach of users that are not expert in
the system's internals and/or specific features.

\section{Concluding remarks}\label{sec:conclusions}
\nopagebreak
Much effort has been spent in the last 20 years
by the ASP community, and outstanding results have
been achieved since the first seminal papers; ASP
and ASP system can be nowadays profitably exploited in many application settings,
not only thanks to the declarative and expressive power of the formalism,
but also thanks to continuously improving performances.
Even a two-year horizon indicates that things are getting better and better.
In addition, it is interesting to note
that, despite the difference between
the specifically tailored solutions and the one with factory settings
being significant, the current state-of-the-art ASP implementations
are offering a good experience to application developers,
given the nice declarative approach of the formalism
and the mature, robust, and currently well-performing available systems.

Nevertheless, there is still much room for improvements. For instance,
the comparison with some ad-hoc solvers confirmed that the performance is
not a tout-court weak point anymore, but the gap with respect to some others
suggests that they might be further improved. The main issue, however,
still remains the lack of a sufficiently broad and standard language.

\paragraph{Acknowledgments.}\label{sec:ack}

All of us feel honored by the invitation of the Universit\`{a} della Calabria
as host institution on the part of the ASP Competition Steering Committee.
We want to thank all the members of the Computer Science Group at the Dipartimento di Matematica of
Universit\`{a} della Calabria for their invaluable collaboration, which made this event possible, and especially: Mario Alviano,
Onofrio Febbraro, Maria Carmela Santoro and Marco Sirianni. We thank Nicola Leone, as the Director of the Dipartimento di Matematica of Universit\`{a} della Calabria, who provided us with all the means, in the form of human and technical resources, and animated earlier discussions we carried out together.
A special thanks goes to all the members of the ASP and CLP communities which authored problem domains, and to participating teams, whose continuous feedback significantly helped in improving competition rules and benchmark specifications.
We thank Graeme Gange and Peter Stuckey for providing us ad-hoc solutions for the {\sc CrossingMinimization} problem originally benchmarked in \mycite{gang-etal-2010-crossingminimization}.
A special mention goes to Jim Delgrande and Wolfgang Faber, for their support as LPNMR-11 conference chairs and editors of the proceedings featuring the preliminary report of the Competition.
We wish also to thank the anonymous reviewers for their fruitful comments and suggestions, that significantly helped to improve the work.

\newcommand{\SortNoOp}[1]{}

\newpage
\appendix

\section{\aspcore Syntax and Semantics overview}\label{app:aspcoreformat}
\nopagebreak
In the following, an overview of both the main constructs and the semantics of the \aspcore language is reported.
The full \aspcore language specification can be found in \mycite{languageformat}.

\subsection{\aspcore Syntax}\label{subapp:aspcoreXyntax}
\nopagebreak
For the sake of readability, the language specification is hereafter given
according to the traditional mathematical notation.
A lexical matching table from the following notation to the actual raw input format
prescribed for participants is provided in \mycite{languageformat}.

\paragraph{Terms, constants, variables.} Terms are either
{\em constants} or {\em variables}. Constants can be either
{\em symbolic constants} (strings starting with lower case letter),
{\em strings} (quoted sequences of characters), or integers.
Variables are denoted as strings starting with an upper case letter.
As a syntactic shortcut, the special variable \quo{$\_$} is a placeholder for
a fresh variable name in the context at hand.

\paragraph{Atoms and Literals.}
An atom is of the form $p(X_1,\dots,X_n)$, where $p$ is a {\em predicate name},
$X_1, \dots, X_n$ are {\em terms},
and $n$ ($n \geq 0$) is the fixed arity associated to $p$%
\footnote{The atom referring to a predicate $p$ of arity 0,
can be stated either in the form $p()$ or $p$.}.
A classical literal is either of the form $a$ ({\em positive classical literal})
or $-a$ ({\em negative classical literal}), for $a$ being an {\em atom}.
A {\em naf-literal} is either a {\em positive naf-literal} $a$ or
a {\em negative naf-literal} $\naf a$, for $a$ being a classical literal.

\paragraph{Rules.} An \aspcore program $P$ is a
a finite set of {\em rules}.
A rule $r$ is of the form
\[
a_1 \vee \dots \vee a_n \leftarrow b_1, \dots , b_k, \naf n_1, ..., \naf n_m.
\]
where $n,k,m \geq 0$, and at least one of $n$,$k$ is greater than $0$;
$a_1,\dots, a_n$, $b_1,\dots,b_k$, and $n_1,\dots,n_m$ are {\em classical literals}.
$a_1 \vee \dots \vee\, a_n$ constitutes the {\em head} of $r$, whereas
$b_1, \dots , b_k, \naf n_1, ..., \naf n_m$ is the {\em body} of $r$.
As usual, whenever $k=m=0$, we omit the \quo{$\leftarrow$} sign.
$r$ is a {\em fact} if $n=1, k=m=0$, while $r$ is a {\em constraint} if $n=0$.

Rules written in \aspcore are assumed to be {\em safe}.
A rule $r$ is safe if all its variables occur in at least one positive naf-literal in the body of $r$.
A program $P$ is safe if all its rules are safe.
A program (a rule, a literal, an atom) is said to be
{\em ground} (or {\em propositional}) if it contains no variables.

\paragraph{Ground Queries.} A program $P$ can be coupled with
a {\em ground query} in the form $q?$, where $q$ is a ground naf-literal.

\subsection{\aspcore Semantics}\label{subapp:aspcoreSemantics}
\nopagebreak
The semantics of \aspcore, based on \mycite{gelf-lifs-91}, exploits
the traditional notion of the Herbrand interpretation.

\paragraph{Herbrand universe.} Given a program $P$,
the {\em Herbrand universe} of $P$, denoted by \UP,
is the set of all constants occurring in $P$.
The {\em Herbrand base} of~$P$, denoted by \BP,
is the set of all ground naf-literals obtainable from the
atoms of $P$ by replacing variables with elements from \UP.

\paragraph{Ground programs and Interpretations.}
A {\em ground instance} of a rule $r$ is obtained replacing
each variable of $r$ by an element from \UP.
Given a program $P$, we define the {\em instantiation (grounding)} \GP
of $P$ as the set of all ground instances of its rules.
Given a ground program $P$, an {\em interpretation} $I$ for $P$ is a subset of \BP.
A {\em consistent} interpretation is such that $\{ a, -a \} \not\subseteq I$ for any ground atom $a$.
In the following, we only deal with consistent interpretations.

\paragraph{Satisfaction.}
A positive naf-literal $l = a$ (resp., a naf-literal $l = \naf\ a$), for predicate atom $a$,
is true with respect to an interpretation $I$ if $a \in I$ (resp., $a \notin I$); it is false otherwise.
Given a ground rule $r$, we say that $r$ is satisfied with respect to an interpretation $I$
if some atom appearing in the head of $r$ is true with respect to $I$
or some naf-literal appearing in the body of $r$ is false with respect to $I$.

\paragraph{Models.}
Given a ground program $P$,
we say that a consistent interpretation $I$ is a {\em model} \ of $P$
iff all rules in \GP are satisfied w.r.t. $I$.
A model $M$ is {\em minimal}
if there is no model $N$ for $P$ such that $N \subset M$.

\paragraph{Gelfond-Lifschitz reduct and Answer Sets.}
The {\em Gelfond-Lifschitz reduct}~\mycite{gelf-lifs-91} of a program $P$
with respect to an interpretation $I$ is the positive ground program $P^I$
obtained from $\GP$ by:
$(i)$ deleting all rules with a negative naf-literal false w.r.t. $I$;
$(ii)$ deleting all negative naf-literals from the remaining rules.
$I\subseteq \BP$ is an {\em answer set} for $P$ iff $I$ is a minimal model for $P^I$.
The set of all answer sets for $P$ is denoted by $AS(P)$.

\paragraph{Semantics of ground queries.}
Let $P$ be a program and $q?$ be a query,
$q?$ is {\em true} iff for all $A \in AS(P)$ it holds that $q \in A$.
Basically, the semantics of queries corresponds to {\em cautious reasoning},
since a query is true if the corresponding atom is true in all answer sets of $P$.

\section{Detailed Competition Settings}\label{app:competitionsettings}
\nopagebreak
\begin{center}
\begin{figure}[b]
 \centering
 \vspace*{3ex}
 \subfigure[$\solver\ Track$]{
  \includegraphics[width=0.44\textwidth]{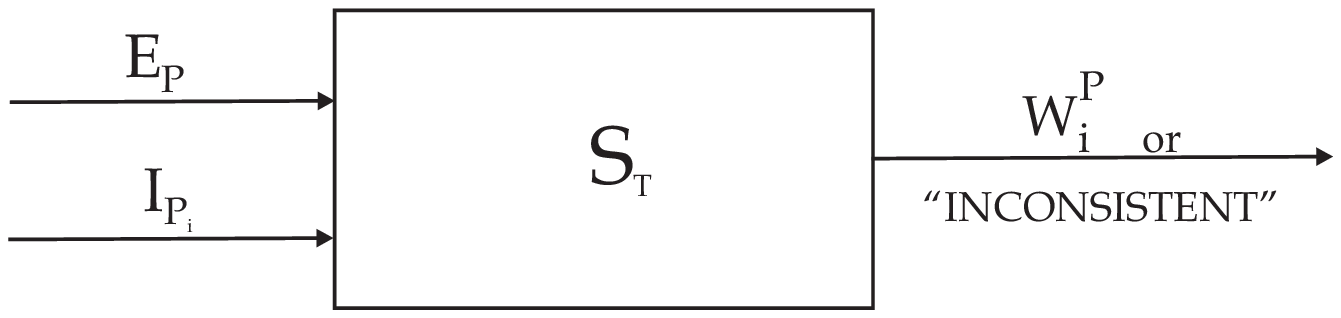}}
 \hfill \hspace{-6ex}
 \subfigure[$\team\ Track$]{
  \includegraphics[width=0.44\textwidth]{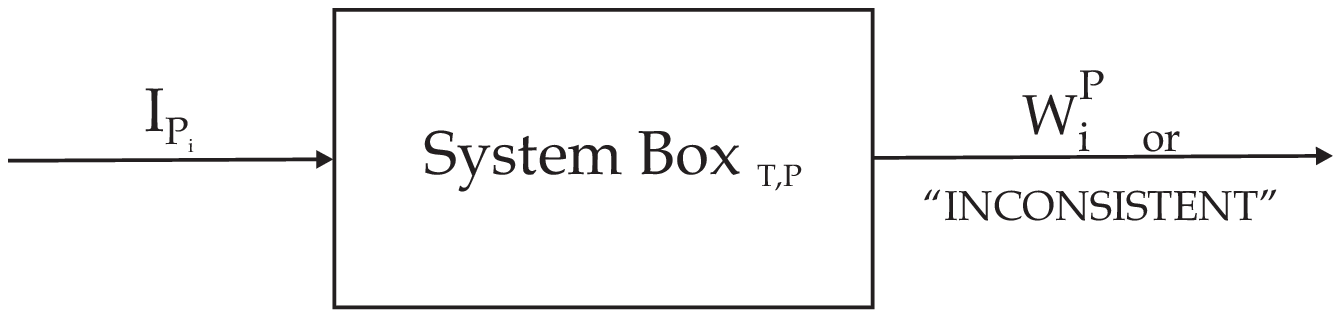}}
 \vspace*{-1ex}
 \caption{\label{fig:competitionSettings} Competition Setting}
\end{figure}
\end{center}

The competition settings for the two tracks are
depicted in Figure \ref{fig:competitionSettings}.
The problems collected into the official problem suite
(see \ref{app:benchmarks}), were grouped into
two different suites, one per each track.
The problems belonging to the \solver Track suite were nearly
a proper subset of the ones featured in the \team Track.
In both tracks, for each problem $P$, a number of instances
$I_{P_1},\dots,I_{P_N}$ were selected.%
\footnote{For problems appearing in both tracks, the instances selected for the \team
Track and those selected for the \solver Track were not necessarily the same.}
For any problem $P$ included into the \solver Track a corresponding fixed
declarative specification, written in \aspcore, $E_P$ was also given.

A team $T$ participating in the \solver Track had
to provide a unique executable system $S_T$.
A team participating in the \team Track, instead, had to produce
a possibly-different execution bundle $SystemBox_{T,P}$
for any problem $P$ in the \team Track suite.
For each problem $P$, the participants were fed iteratively
with all instances $I_{P_i}$ of $P$ (in the case of the \solver Track
each instance was fed together with the corresponding problem encoding $E_P$).

The submitted executables were challenged to produce either
a {\em witness} solution, denoted by $W_i^P$, or to report
that no solution exists within a predefined amount of allowed time.
The expected output format that is determined by the type of the
problem (search, query, optimization) is reported in \mycite{languageformat}.
Participants were made aware, fairly in advance,
of fixed encodings (in the case of the \solver Track), while they were provided only a
small set of corresponding training instances.
Official instances were kept secret until the actual start of the competition.
Scores were awarded according to the competition scoring system
(see Section \ref{sec:compsettings} and \ref{app:detailedscoring}).

\paragraph{\bf Definition of \quo{syntactic special purpose technique}.}
    The committee classified as forbidden in the \solver Track: the switch of internal solver options depending either
    on command-line filenames, predicate and variable names, and ``signature'' techniques aimed
    at recognizing a particular benchmark problem, such as counting the number of rules,
    constraints, predicates and atoms in a given encoding.
    In order to discourage the adoption of forbidden techniques,
    the organizing committee reserved the right to introduce syntactic means
    for scrambling program encodings, such as file, predicate and variable random renaming.
    Furthermore, the committee reserved the right to replace official program encodings arbitrarily
    with equivalent syntactically-changed versions.

    It is worth noting that, on the other hand, the semantic recognition of the program
    structure was allowed, and even encouraged.
    Allowed semantic recognition techniques explicitly included:
    {\em (i)} recognition of the class the problem encoding belongs to
    (e.g., stratified, positive, etc.), with possible consequent
    switch-on of on-purpose evaluation techniques;
    {\em (ii)} recognition of general rule and program structures
    (e.g., common un-stratified even and odd-cycles, common join patterns within a rule body, etc.),
    provided that these techniques were general and not specific of a given problem selected
    for the competition.

\paragraph{Detailed Software and Hardware Settings.}
\label{app:competition-run}
The Competition took place on a battery of four servers, featuring a 4-core Intel Xeon CPU X3430 running at 2.4 Ghz, with 4 GiB of physical RAM and PAE enabled.

The operating system of choice was Linux Debian Lenny (32bit), equipped with the C/C++ compiler GCC 4.3 and common scripting/development tools. Competitors were allowed to install their own compilers/libraries in local home directories, and to prepare system binaries for the specific Competition hardware settings.
All the systems were benchmarked with just one out of four processors enabled, with the exception
of the parallel solver \clasp-{\tt \small mt} that could exploit all the available core/processors.
Each process spawned by a participant system had access to the usual Linux process memory
space (slightly less than 3GiB user space + 1GiB kernel space). The total memory allocated
by all the child processes created was however constrained to a total of 3 GiB (1 GiB = $2^{30}$ bytes).
The memory footprint of participant systems was controlled by using the Benchmark
Tool Run.\footnote{\url{http://fmv.jku.at/run/}.}
This tool is not able to detect short memory spikes (within 100 milliseconds) or, in some corner
cases, memory overflow is detected with short delay: however, we pragmatically assumed the tool as
the official reference.

\paragraph{Detection of Incorrect Answers.}
\label{subapp:incorrect}
Each benchmark domain $P$ was equipped with a checker program $C_P$ taking as input values a witness $A$ and an instance $I$, and answering \quo{true} in case $A$ is a valid witness for $I$ w.r.t problem $P$.
The collection of checkers underwent thorough assessment and then was pragmatically assumed to be correct.

Suppose that a system \systemS is faulty for instance $I$ of problem $P$; then, there were two possible scenarios
in which incorrect answers needed detection and subsequent disqualification for a given system:
\begin{itemize}
\item
\systemS produced an answer $A$, and $A$ was not a correct solution
 (either because $I$ was actually unsatisfiable or $A$ was wrong at all).
 This scenario was detected by checking the output of $C_P(A,I)$;
\item
\systemS answered that the instance was not satisfiable, but actually $I$ had some witness.
  In this case, we checked whether a second system $\systemS'$ produced a solution $A'$ for which $C_P(A',I)$ was true.
\end{itemize}

Concerning optimization problems, checkers produced also the cost $C$ of the given witness. This latter value was considered when computing scores and for assessing answers of systems.
Note that cases of general failure (e.g. out of memory, other abrupt system failures)
were not subject of disqualification on a given benchmark.
As a last remark, note that in the setting of the \system Track, where problem encodings were fixed, a single
stability checker for answer sets could replace our collection of checkers. We preferred to exploit already available
checker modules, which were also used for assessing the correctness of fixed official encodings set for the \system Track.
This enabled us to detect some early errors in fixed encodings: however, our lesson learned suggests that a general stability
checker should be placed side-by-side to specific benchmark checkers.

\paragraph{Other settings.}
The committee kept its neutral position and did not disclose any material submitted by participants
until the end of the competition: however, participants were allowed to share their own work willingly at any moment.
The above choice was taken in order to prefer scientific collaboration between teams over a strict competitive setting.
All participants were asked to agree that any kind of submitted material (system binaries, scripts, problems encodings, etc.)
was to be made public after the competition, so to guarantee transparency and reproducibility.
None of the members of the organizing committee submitted a system to the Competition,
in order to play the role of neutral referee properly and guarantee an unbiased benchmark selection and rule definition process.

\section{Detailed scoring regulations}\label{app:detailedscoring}
\nopagebreak

\subsection{Principles}\label{subapp:guidelines}
\nopagebreak
The main factors that were taken into account in the scoring framework are illustrated next.

\begin{enumerate}
\item
Benchmarks with many instances should not dominate the overall score of a category.
Thus, the overall score for a given problem $P$ was normalized
with respect to the number $N$ of selected instances for $P$.

\item
Nonsound solvers and encodings were strongly discouraged.
Thus, if system $S$ produced an incorrect answer for an instance of a problem $P$
then $S$ is disqualified from $P$ and the {\em overall score} achieved by $S$ for problem $P$
is invalidated (i.e., is set to zero).

\item
A system managing to solve a given problem instance sets a clear gap over all systems not able to do so.
Thus, a flat reward for each instance $I$ of a problem $P$ was given
to a system $S$ that correctly solved $I$ within the allotted time.

\item
Concerning time performance, human beings are generally
more receptive to the logarithm of the changes of a value,
rather than to the changes themselves; this is especially the case
when considering evaluation times. Indeed, different systems with time performances
being in the same order of magnitude are perceived as comparatively similar, in
terms of both raw time performance and quality; furthermore, a system is
generally perceived as clearly \emph{fast}, when its solving times
are orders of magnitude below the maximum allowed time.
Keeping this in mind, and analogously to what has been done in SAT competitions,%
\footnote{See, for instance, the log based scoring formulas
at \url{http://www.satcompetition.org/2009/spec2009.html}.}
a logarithmically weighted bonus was awarded to faster systems depending
on the time needed for solving each instance.

\item
In the case of optimization problems, scoring should depend also
on the quality of the provided solution. Thus, bonus points were rewarded
to systems able to find better solutions. Also, we wanted to take
into account the fact that small improvements in the quality of a solution
are usually obtained at the price of much stronger computational efforts:
thus the bonus for a better quality solution has been given
on an exponential weighting basis.
\end{enumerate}

\begin{figure}[b]
\vspace{-0.1cm}
\includegraphics[width=0.9\textwidth]{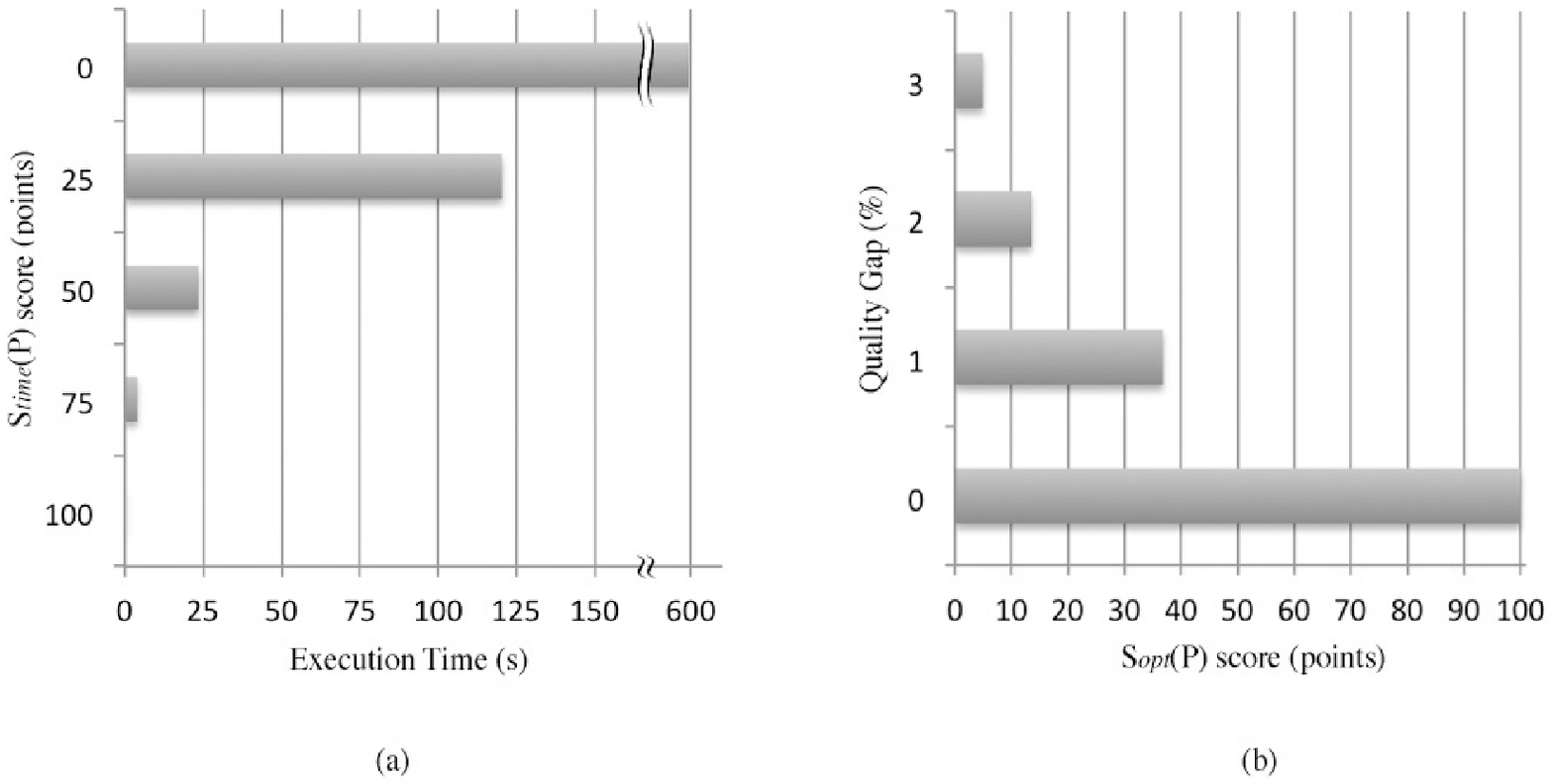}
\vspace{-0.3cm}
\caption{Scoring Functions Exemplified (one instance, 100 pts max, $t_{out} = 600$). \label{fig:stime}}
\vspace{-0.5cm}
\end{figure}

\subsection{Scoring Rules}\label{subapp:scoring-framework}
\nopagebreak
The final score obtained by a system \systemS in a track \trackT consisted of the sum over
the scores obtained by \systemS in all benchmarks selected for \trackT.
In particular, a system could get a maximum of \pPointsT points for each given benchmark problem $P$ considered for \trackT.
The overall score of a system on a problem $P$ counting $N$ instances, hereafter denoted by $S(P)$,
was computed according to the following formulas that depend on whether $P$ is a search, query or optimization problem.

\paragraph{Wrong Answers.} In the case where \systemS produced an output detected as incorrect%
\footnote{Incorrect answers were determined as specified in \ref{subapp:incorrect}}
for at least one instance of $P$, then \systemS was disqualified from $P$
and $S(P)$ was set to zero (i.e., $S(P)=0$ in case of incorrect output);
otherwise, the following formulas were applied for computing $S(P)$.

\paragraph{Search and Query Problems.}
\label{subapp:search-score}
In case of both search and query problems the score $S(P)$ was computed by the sum
 $$ S(P) = S_{solve}(P)+ S_{time}(P) $$

\noindent where $S_{solve}$ and $S_{time}(P)$  take into account the number of instances solved by \systemS in $P$
and the corresponding running times, respectively; in particular
 $$S_{solve}(P) = \alpha \frac {N_S}{N} \ ;\ \ \  S_{time}(P) = \frac{100 - \alpha}{N} \sum_{i=1}^{N}{ \left(1 - \left( \frac{\log ( t_i + 1 ) }{\log (t_{out} + 1 )}\right) \right) } $$

\noindent for ${N_S}$ being the number of instances solved by $P$ within the time limit,
$t_{out}$ is the maximum allowed time,
$t_{i}$ the time spent by $S$ while solving instance $i$, and $\alpha$ a percentage factor
balancing the impact of $S_{solve}(P)$ and $S_{time}(P)$ on the overall score.
Both $S_{solve}(P)$ and $S_{time}(P)$ were rounded to the nearest integer.

Note that $S_{time}(P)$ was specified in order to take into account the ``perceived''
performance of a system (as discussed in \ref{subapp:guidelines}).
Figure \ref{fig:stime}(a) gives an intuitive idea about how $S_{time}$ distributes a maximum score of 100 points
considering a single instance and $t_{out} = 600$. Note that half of the maximum score (50 points)
is given to performance below 24 seconds about, and significant differences
in scoring correspond to differences of orders of magnitude in time performance.

\paragraph{Optimization Problems.}\label{subapp:optimization-score}
As in the previous edition, the score of a system \systemS in the case of optimization problems depends on
whether \systemS was able to find a solution or not, and in the former case,
the score depends on the quality of the given solutions. In addition,
as in the case of decision problems, time performance is taken into account.
We assumed the cost function associated with optimization problems must be minimized
(the lower, the better), and it had $0$ as its lowest bound.

The overall score of a system for an optimization problem $P$ is given by the sum
 $$S(P) = S_{opt}(P)+S_{time}(P) $$

\noindent where $S_{time}(P)$ is defined as for search problems,
and $S_{opt}(P)$ takes into account the quality of the solution found.
In particular, for each problem $P$, system \systemS is rewarded of a number of points defined as
$$S_{opt}(P) = \alpha \cdot \sum_{i=1}^{N} S_{opt}^{i}$$

\noindent where, as before, $\alpha$ is a percentage factor balancing the impact of $S_{opt}(P)$ and $S_{time}(P)$ on the overall score, and $S_{opt}^{i}$ is computed by properly summing, for each instance $i$ of $P$, one or more of these rewards:
\begin{enumerate}
\item $\frac {1}{N}$ points, if the system correctly recognizes an unsatisfiable instance;
\item $\frac {1}{4N}$ points, if the system produces a correct witness;
\item $\frac {1}{4N}$ points, if the system correctly recognizes an optimum solution and outputs it;
\item $\frac {1}{2N} \cdot  e^{M - Q}$ points, where $Q$ denotes the quality of the solution produced by the system
and $M$ denotes the quality of the best answer produced by any system for the current instance, for $M$ conventionally set to $100$, and $Q$ normalized accordingly.
\end{enumerate}

Taking into account that an incorrect answer causes the whole benchmark
to pay no points, three scenarios may come out: timeout,
unsatisfiable instance, or solution produced.
Note thus that points of groups (1), (2) and (3-4-5) cannot be rewarded for the same instance.

The intuitive impact of the above ``quality'' score $S_{opt}(P)$ can be seen in Figure \ref{fig:stime}(b),
in which the quality of a given solution, expressed in percentage distance from the optimal solution,
is associated with the corresponding value of $S_{opt}$ (suppose a maximum of 100 points, $\alpha = 100$, and one single instance per benchmark).
Note that a system producing a solution with a quality gap of 1$\%$ with respect to the best solution
gets only 35 points (over 100) and the quality score quota rapidly
decreases (it is basically $0$ for quality gap $> 4\%$), so that
small differences in the quality of a solution determine a strong difference in scoring
according to considerations made in \ref{subapp:guidelines}.

In the present competition, for each problem domain
we set  $t_{out} = 600$ seconds and $\alpha = 50$;
$N$ has been set to 10 for the \solver Track, while it varied from problem
to problem for the \team Track, reaching up to 15 instances per
single benchmark problem.

\section{Benchmark Suite}\label{app:benchmarks}
\nopagebreak
Benchmark problems were collected, selected and refined
during the {\em Call for Problems} stage.
The whole procedure led to the selection of 35 problems, which constituted the \team Track problem suite. Taking into account what already discussed in Section \ref{sec:competition-format},
twenty problems out of the ones constituting the \team Track
were selected for composing the \solver Track suite: these had a natural and declarative \aspcore encoding. Benchmark problems were classified according to their type into three categories:
{\em Search problems}, requiring to find a solution (a {\em witness}) for
the problem instance at hand, or to notify the non-existence of a solution; {\em Query problems},
consisting in checking whether a ground fact is contained in all the witnesses
of the problem instance at hand (same as performing cautious reasoning on a given logic program); and,
{\em Optimization problems}, i.e. a search problem in which a cost function associated to witnesses
had to be minimized. The \solver Track did not contain optimization problems.

\medskip
Problems were further classified according to their computational complexity
in three categories:%
\footnote{The reader can refer to \mycite{papa-94} for the definition
of basic computational classes herein mentioned.}
{\em Polynomial, NP} and {\em Beyond NP} problems, these latter with the two
subcategories composed of $\Sigma^P_2$ problems and optimization problems.
In the following, we break down the benchmark suite according
complexity categories and discuss some interesting aspects.
The complete list of problems included in the competition benchmark suite,
together with detailed problem descriptions, and full benchmark data,
is available on the Competition web site \mycite{aspcomp2011-web}.

\paragraph{\bf Polynomial Problems.}\label{subsec:poly}
We classified in this category problems which are known to be solvable in polynomial time in
the size of the input data (data complexity).
In the Competition suite such problems were usually characterized by the huge size of instance data
and, thus, they were a natural test-bench for the impact of memory consumption on performance.
It is worth disclaiming that the competition aim was not to compare ASP systems
against technologies (database etc.) better tailored to solving this category of problems;
nonetheless, several practical real-world applications,
which competitors should be able to cope with, fall into this category.
Note also that polynomial problems are usually entirely solved by participants' grounder modules,
with little or no effort required by subsequent solving stages: indeed, grounders are the technology
that mainly underwent assessment while dealing with polynomial problems.
There were seven polynomial problems included in the benchmark suite, six of which
were selected for the \solver Track suite.

Four of the above six were specified in a fragment of \aspcore (i.e., stratified logic programs)
with polynomial data complexity; a notable exception was made by the problems
{\sc StableMarriage} and {\sc PartnerUnitsPolynomial}
--which are also known to be solvable in polynomial time
\mycite{gusf-etal-1989-stable-marriage,falk-hase-2010-ecai}--
for which we chose their natural declarative encoding, making usage of disjunction.
Note that, in the case of these last two problems,
the \quo{combined} ability of grounder and propositional solver modules was tested.
The aim was to measure whether, and to what extent, a participant system
could be able to converge on a polynomial evaluation strategy when
fed with such a natural encoding.

As further remark, note that the polynomial problem {\sc Reachability}
was expressed in terms of a query problem,
in which it was asked whether two given nodes were reachable in a given graph:
this is a typical setting in which one can test systems on their search
space tailoring techniques (such as {\em magic sets})~\mycite{banc-etal-1986}.
The polynomial problem participating to the \team Track only was {\sc CompanyControls},
given its natural modeling in term of a logic program with
aggregates~\mycite{fabe-etal-2004-jelia}, these latter not included in the \aspcore specifications.

\paragraph{\bf NP Problems.}\label{subsec:NP}

We classified in this category NP-complete problems or, more precisely, their
corresponding FNP versions.
These problems constituted the \quo{core} category, in which to test
the attitude of a system in efficiently dealing with problems expressed with
the \quo{Guess and Check} methodology \mycite{leon-etal-2002-dlv}.

Among the selected NP problems there were ten puzzle problems, six of which inspired
by or taken from planning domains; two classical graph problems;
six, both temporal and spatial, resource allocation problems; and,
three problems related to applicative and academic settings, namely:
{\sc Weight-AssignmentTree} \mycite{ullm-etal-00} which was concerned with
the problem of finding the best join ordering in a conjunctive query;
{\sc ReverseFolding} which was aimed at mimicking the protein folding problem in a simplified setting \mycite{dovier-2011};
and,
{\sc MultiContextSystemQuerying}, the unique problem considered in the \solver Track only,
which was a query problem originating from reasoning tasks in Multi-Context Systems~\mycite{daot-etal-KR2010-distributed}.
Notably, this latter problem had an \aspcore encoding producing several logic submodules,
each of which with independent answer sets. The ability to handle both cross-products
of answer sets and early constraint firing efficiently were herein assessed.

\paragraph{\bf Beyond NP/$\Sigma^P_2$.}
The category consisted of problems whose decision version was $\Sigma^P_2$-complete.
Since a significant fraction of current ASP systems cannot properly handle this class of problems,
only two benchmarks were selected, namely {\sc StrategicCompanies} and {\sc MinimalDiagnosis}.
The former is a traditional $\Sigma^P_2$ problem coming from \mycite{cado-etal-97},
while the latter originates from an application in molecular biology \mycite{gebs-etal-2011-tplp-bio}.
As far as the \solver Track is concerned, $\Sigma^P_2$ problems have an ASP encoding
making unrestricted usage of disjunction in rule heads.

\paragraph{{\bf Beyond NP}/Optimization.}
These are all the problems with an explicit formulation
given in terms of a cost function with respect to each witness has to be minimized.
The above categorization does not imply a given problem
stays outside $(F)\Sigma^P_2$ , although this has been generally
the case for this edition of the competition.
The selected problems were of heterogenous provenance, including
classic graph problems and sequential optimization planning problems.
No benchmark from this category was present in the \solver Track benchmark suite.

\newpage
\section{System Versions}\label{app:systemVersions}
As described in Section~\ref{sec:participants}, the participants submitted
original systems and solution bundles possibly relying on different (sub)systems.

In some cases, systems were provided as custom versions compiled on purpose for
the Competition; in some other cases, the executables came from the official release
sources, but have been built on the competition machines, and hence might differ
from the ones officially distributed. We explicitly report here the exact versions,
whenever applicable, if explicitly stated by the participants; it is worth remembering
that, for the sake of reproducibility, all systems and solution bundles, together with
encodings, instances, scripts, and everything else needed for the actual re-execution of
the competition, is available on on the competition web site \cite{aspcomp2011-web},
where more details on systems and teams can be found, as well.

\begin{small}
\begin{center}
\begin{tabular}{ll}\label{table:systemVersions}
\textbf{System}                       & \textbf{Related Systems/Subsystems}        \\
$\bullet$ \clasp\ (v 2.0.0-RC2)       & $\bullet$ \Gringo\ (v.3.0.3) \\
$\bullet$ \claspD\ (v 1.1.1)          & $\bullet$ \Gringo\ (v.3.0.3) \\
$\bullet$ \claspfolio\ (v 1.0.0)      & $\bullet$ \Gringo\ (v.3.0.3) \\
$\bullet$ \IDP\ (custom)              & $\bullet$ \Gringo\ (v.3.0.3),\ {\sc MiniSatID}\ (v. 2.5.0) \\
$\bullet$ \cmodels\ (v 3.81)          & $\bullet$ \Gringo\ (v.3.0.3),\ {\sc MiniSat}\ v 2.0-beta \\
$\bullet$ {\sc sup}\ (v 0.4)          & $\bullet$ \Gringo\ (v.3.0.3) \\
$\bullet$ {\sc lp2gminisat},          & $\bullet$ \Gringo\ (v. 3.0.3),\ {\sc Smodels}\ (v. 2.34),\ lpcat (v. 1.18),\ \\
 \ \ \ \ {\sc lp2lminisat},           & \ \ \ \ \ lp2normal (v. 1.11),\ igen (v. 1.7),\ lp2lp2 (v. 1.17),\  \\
 \ \ \ \ {\sc lp2minisat}\            & \ \ \ \ \ {\sc lp2sat}\ (v 1.15),\ {\sc Minisat}\ (v. 1.14),\ interpret (v. 1.7) \\
$\bullet$ {\sc lp2diffz3}             & $\bullet$ \Gringo\ (v. 3.0.3),\ {\sc Smodels}\ (v. 2.34),\ lpcat (v. 1.18),\ \\
 \ \ \ \                              & \ \ \ \ \ l2diff (v.  1.27),\  z3 (v. 2.11),\ interpret (v. 1.7) \\
$\bullet$ {\sc Smodels} (v. 2.34)     & $\bullet$ \Gringo\ (v.3.0.3) \\
 & \\
\textbf{Team}               & \textbf{System/Subsistems exploited}        \\
$\bullet$ \Aclasp           & $\bullet$ \clasp\ (custom),\ \Gringo\ (v.3.0.4) \\
$\bullet$ BPSolver          & $\bullet$ B-Prolog (v. 7.1),\ \\
$\bullet$ EZCSP             & $\bullet$ ezcsp (v. 1.6.20b26),\ \iClingo\ (v. 3.0.3), \clasp,\ \\
 \ \ \ \                    & \ \ \ \ \ ASPM,\ B-Prolog,\ MKAtoms (v. 2.10),\ \Gringo\ (v. 3.0.3) \\
$\bullet$ Fast Downward     & $\bullet$ Fast Downward (custom) \\
$\bullet$ \IDP              & $\bullet$ {\sc Gidl}\ v. 1.6.12,\ {\sc MiniSatID}\ (v. 2.5.0) \\
$\bullet$ Potassco          & $\bullet$ \clasp\ (v 2.0.0-RC2),\ \claspD\ (v 1.1.1),\ \Gringo\ (v.3.0.3),\ \\
 \ \ \ \                    & \ \ \ \ \ \Clingcon\ (v. 0.1.2)
\end{tabular}
\end{center}
\end{small}

\section{Detailed result tables for Section \ref{sec:results}}\label{app:results}
\nopagebreak
We report here detailed figures of the competition.

All the graphs plot a number representing a number of instances (horizontal
axis) against the time (expressed in seconds) needed by each solution bundle
to solve them (vertical axis): the slower a line grows, the more efficient
the corresponding solution bundle performed. Note that not all the participants
solved the same number of instances within the maximum allotted time.

In figure \ref{fig:table-system-totals} and \ref{fig:table-team-totals}
participants are ordered by {\em Final score} (i.e., $\sum_P S(P)$);
for each participant, three rows report for each problem $P$:
$(i)$ the {\em Score},
$(ii)$ the {\em Instance} quota (i.e., $\sum_P S_{solve}(P)$ or $\sum_P S_{opt}(P)$ for optimization problems),
$(iii)$ the {\em Time} quota (i.e., $\sum_P S_{time}(P)$).

For each category, the best performance among official participants is reported
in bold face. In all tables, an asterisk (`$*$') indicates that the system/team has
been disqualified for the corresponding benchmark problem.

\begin{figure}[bh]
 \centering
\vspace*{-1ex}
  \includegraphics[width=0.90\textwidth]{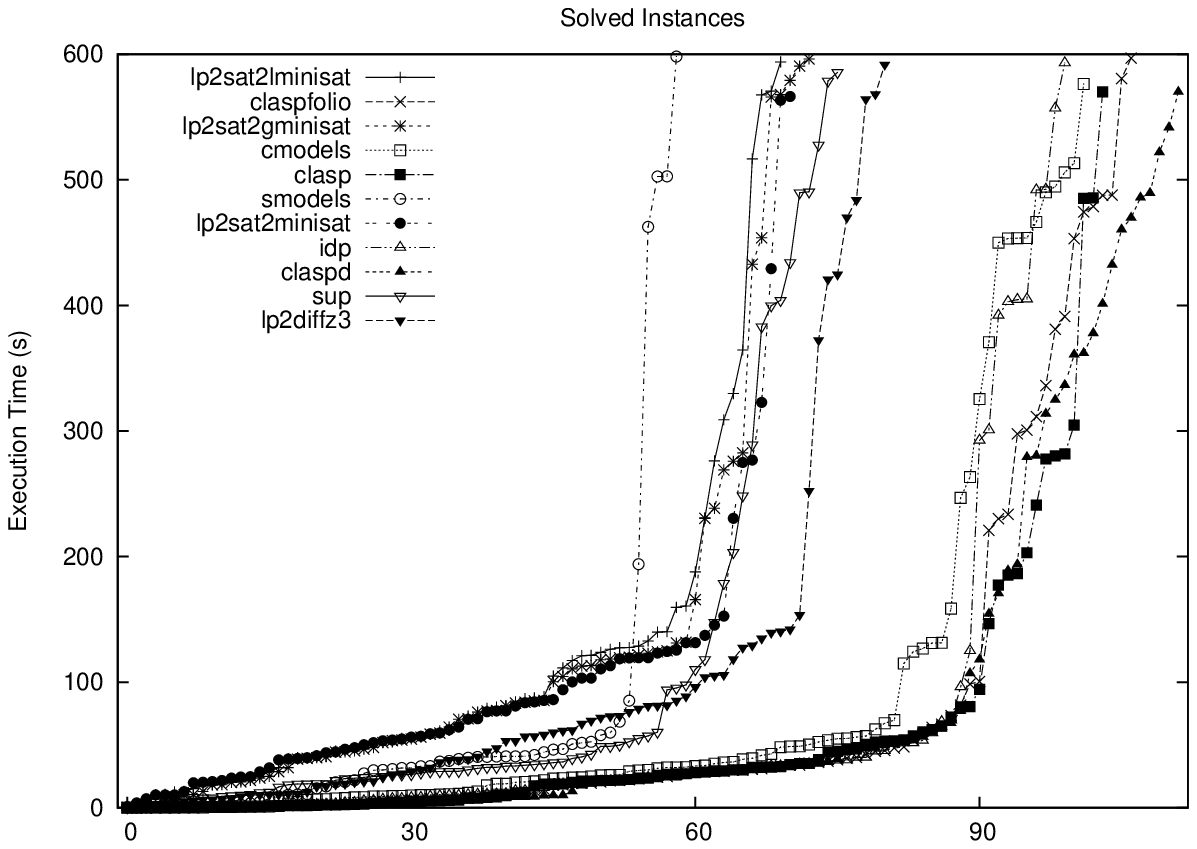}
\vspace*{-1ex}
 \caption{\label{fig:plots-overall-system} \textit{\system Track}: Overall Results [Exec. time (y-axis), Solved Instances (x-axis)]}
\vspace*{-1ex}
\end{figure}

\begin{figure}[bh]
 \centering
\vspace*{-1ex}
  \includegraphics[width=0.90\textwidth]{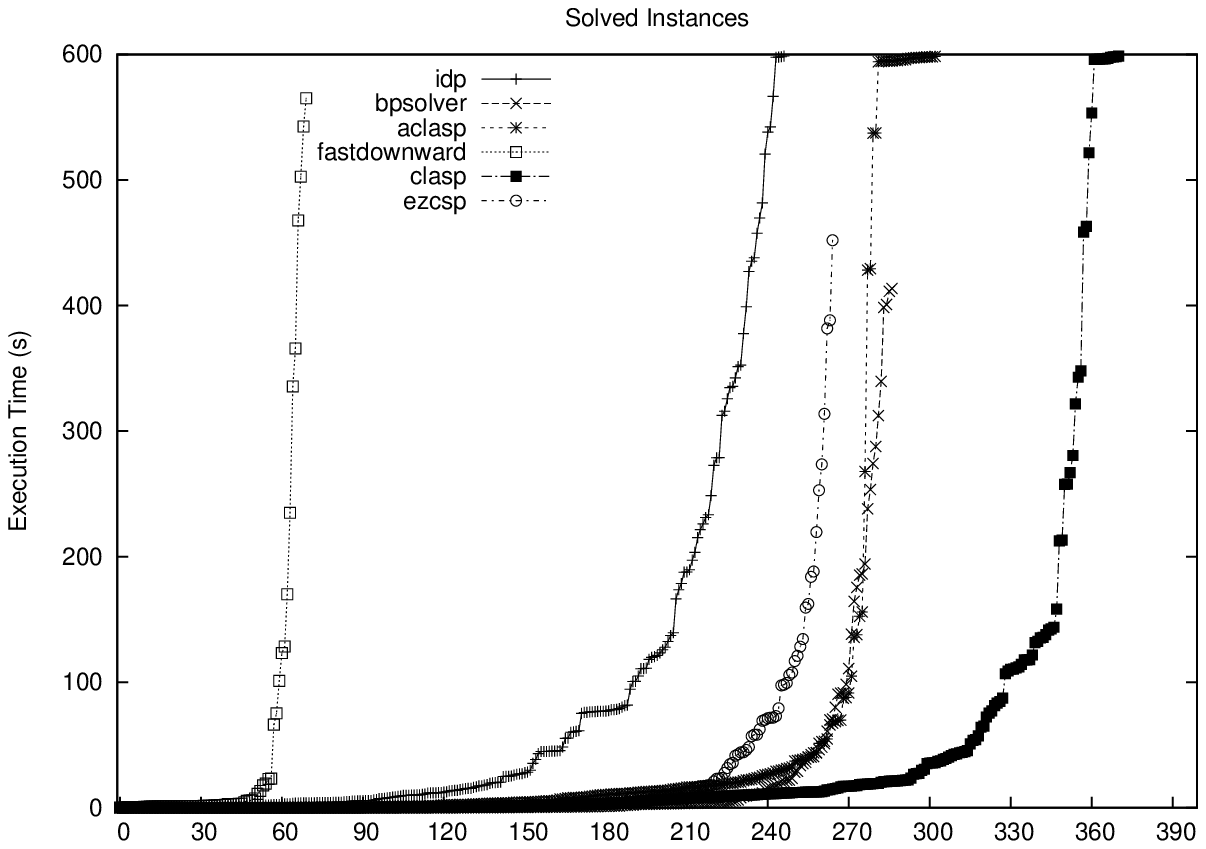}
\vspace*{-1ex}
 \caption{\label{fig:plots-overall-team} \textit{\team Track}: Overall Results [Exec. time (y-axis), Solved Instances (x-axis)]}
\vspace*{-1ex}
\end{figure}

\begin{center}
\begin{figure}[tb]
 \centering
  \includegraphics[width=.999\textwidth]{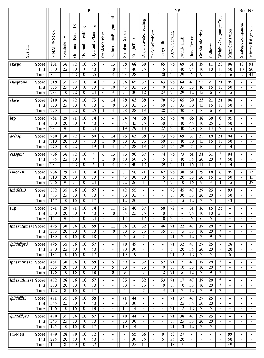}
 \caption{\system Track - Overall Results}\label{fig:table-system-totals}
\end{figure}
\end{center}

\begin{landscape}
\begin{center}
\begin{figure}[!ht]
  \includegraphics[width=1.55\textheight]{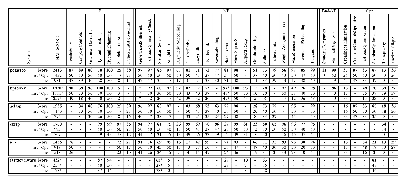}
 \caption{\team Track - Overall Results}\label{fig:table-team-totals}
\end{figure}
\end{center}
\end{landscape}

\begin{figure}[!t]
 \centering
 \subfigure[System Track $P$]{
  \includegraphics[width=0.77\textwidth,viewport=10 0 330 300]{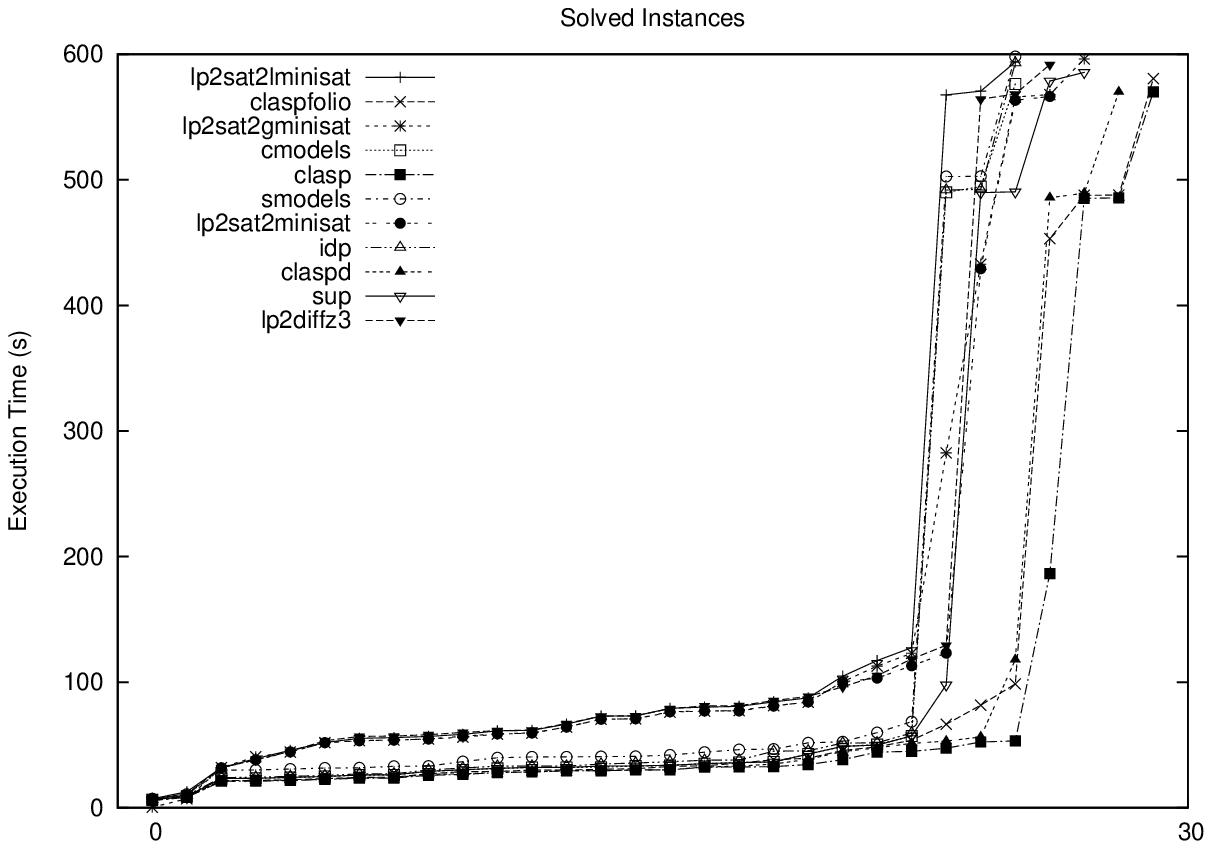}\label{fig:plots-detail-p}}
 \subfigure[Team Track $P$]{
  \includegraphics[width=0.77\textwidth,viewport=10 0 330 300]{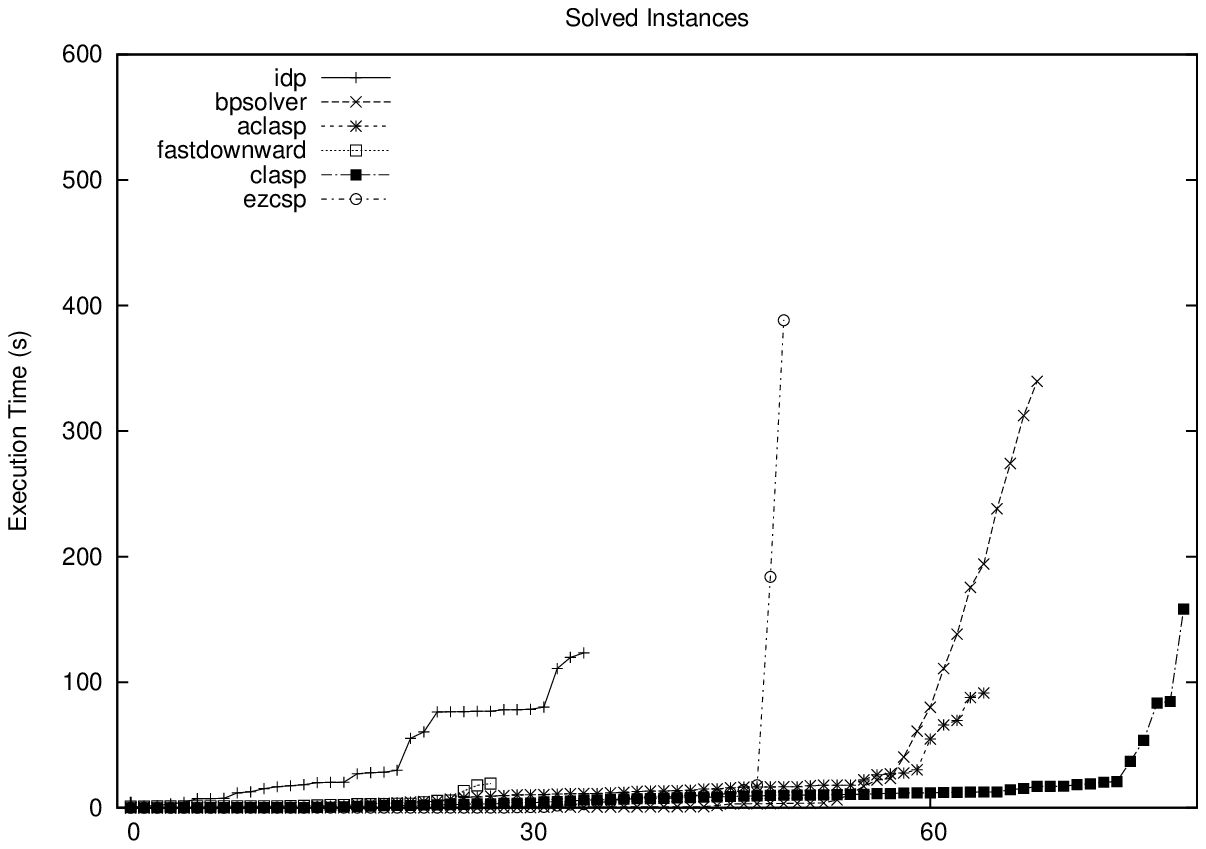}\label{fig:plots-detail-p-team}}\\
   \vspace*{1ex}
 \caption{Results in Detail: Execution time (y-axis), Solved Instances (x-axis). }\label{fig:plots-detail-1}
\end{figure}

\begin{figure}[!t]
 \centering
 \subfigure[System Track $N\!P$]{
  \includegraphics[width=0.77\textwidth,viewport=10 0 330 300]{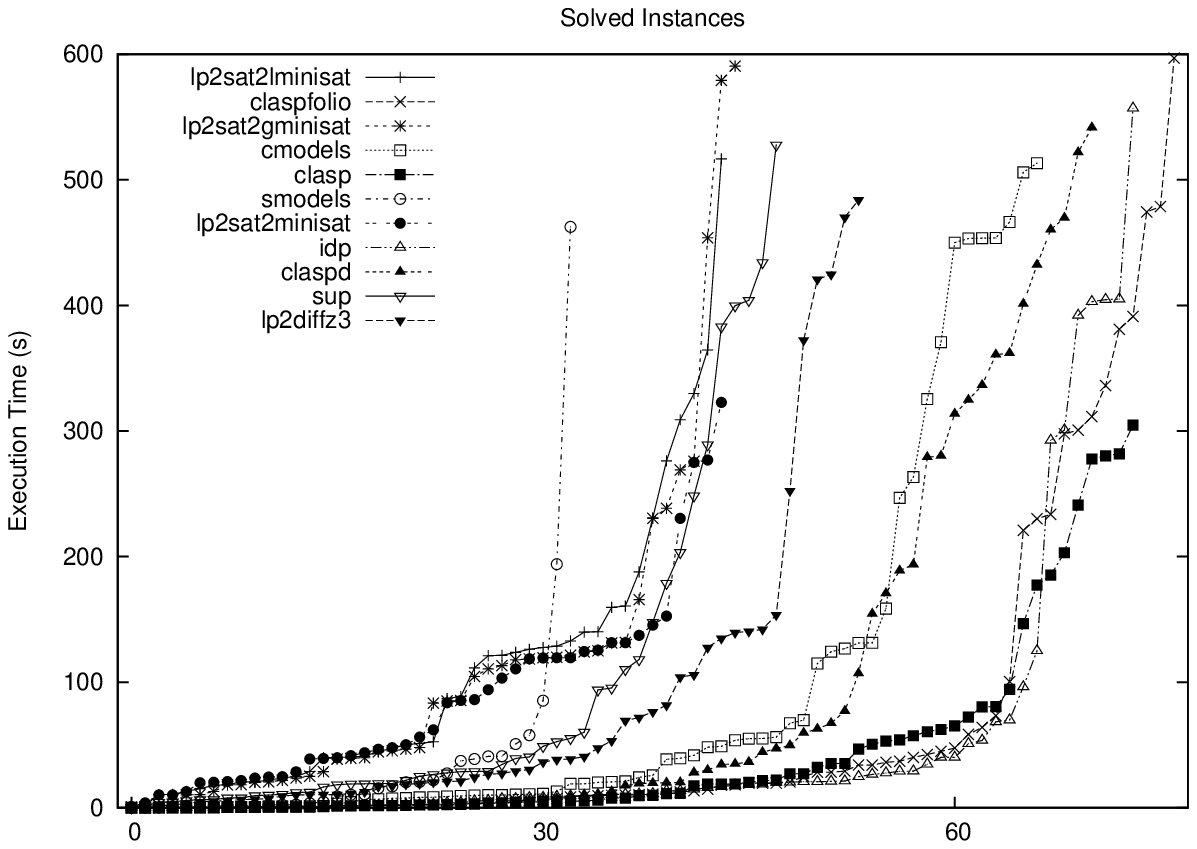}\label{fig:plots-detail-np}}
 \subfigure[Team Track $N\!P$]{
  \includegraphics[width=0.77\textwidth,viewport=10 0 330 300]{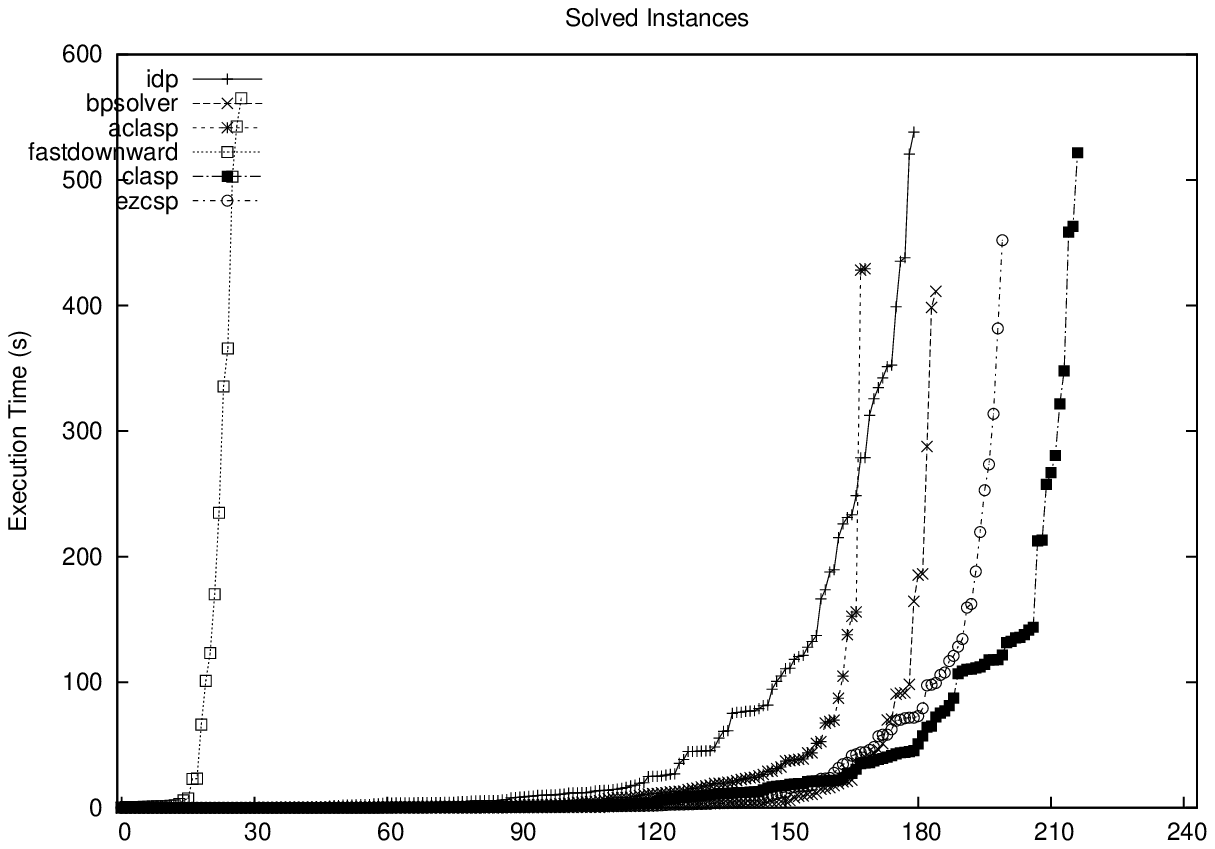}\label{fig:plots-detail-np-team}}\\
   \vspace*{1ex}
 \caption{Results in Detail: Execution time (y-axis), Solved Instances (x-axis).}\label{fig:plots-detail-2}
\end{figure}

\begin{figure}[!t]
 \centering
 \subfigure[System Track Beyond $N\!P$]{
  \includegraphics[width=0.77\textwidth,viewport=10 0 330 300]{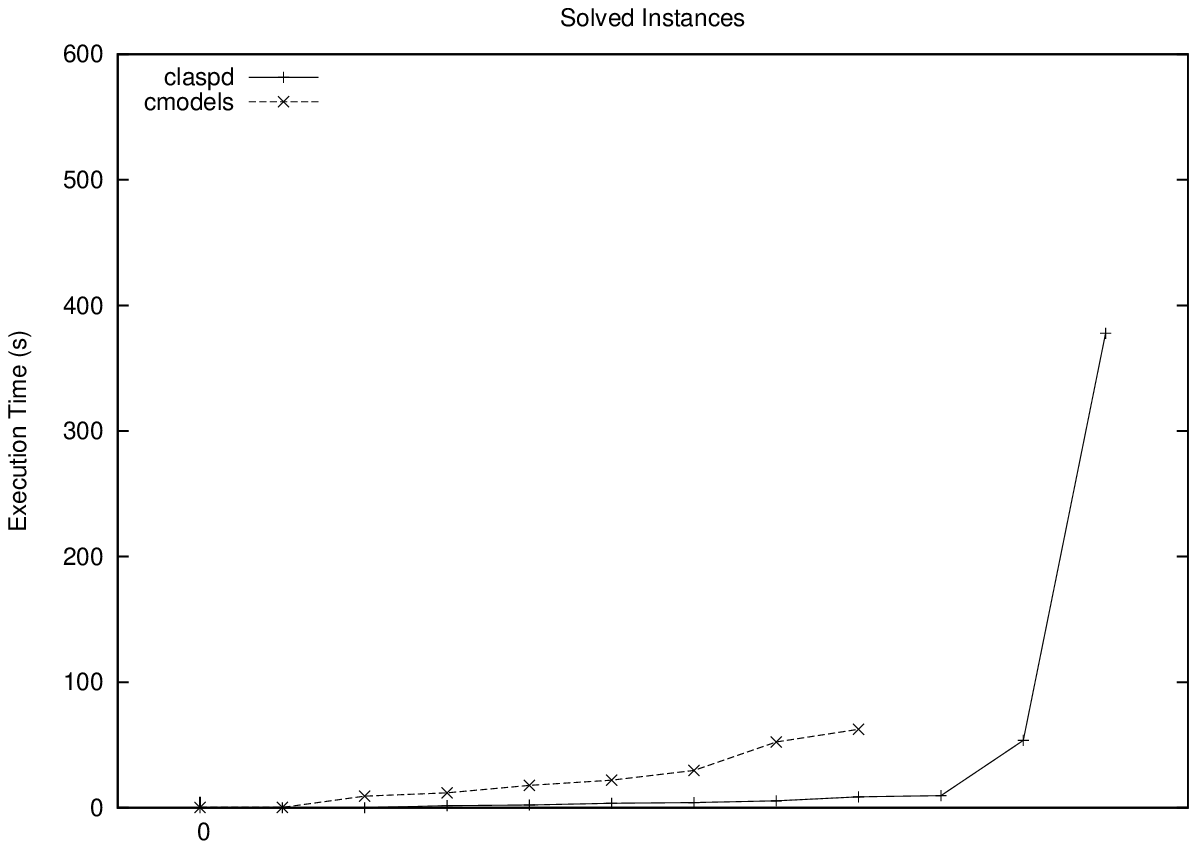}\label{fig:plots-detail-sigmap2}}
 \subfigure[Team Track Beyond $N\!P$]{
  \includegraphics[width=0.77\textwidth,viewport=10 0 330 300]{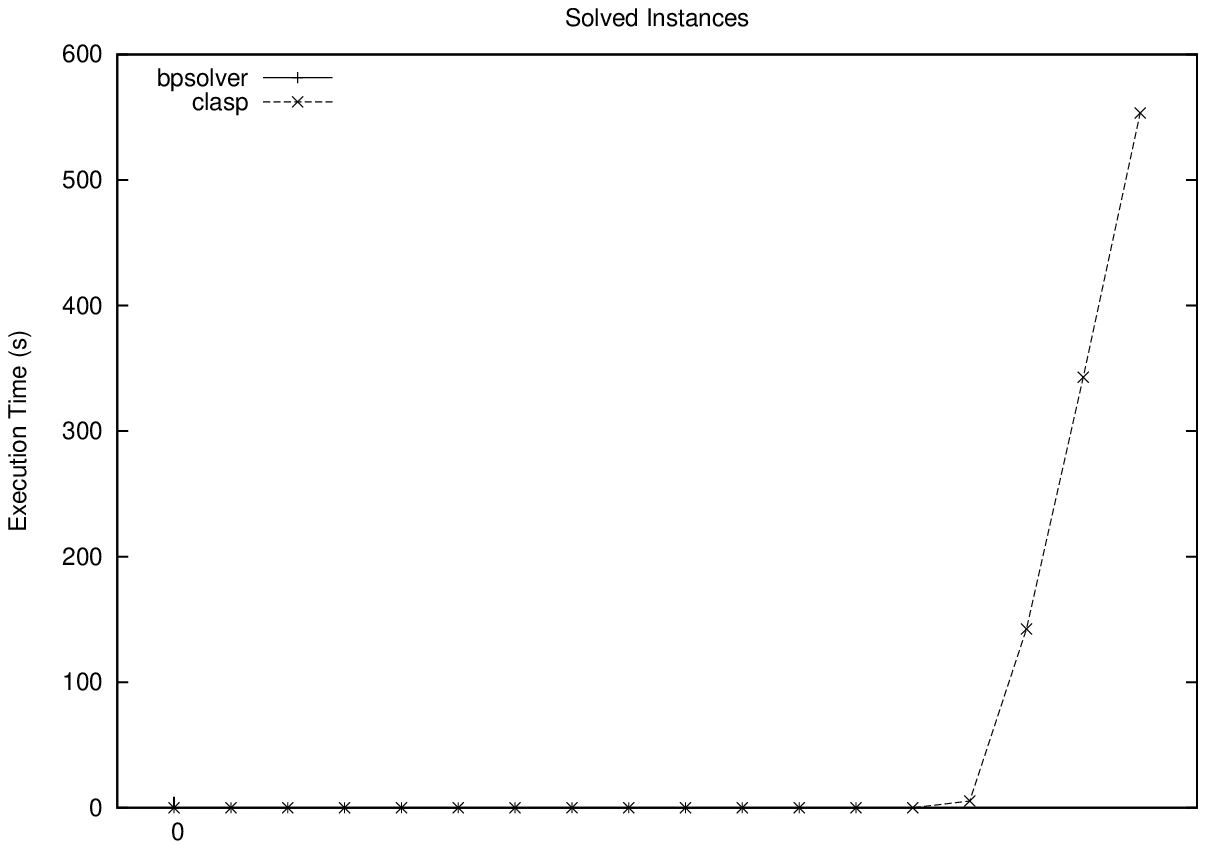}\label{fig:plots-detail-sigmap2-team}}\\
   \vspace*{1ex}
 \caption{Results in Detail: Execution time (y-axis), Solved Instances (x-axis). }\label{fig:plots-detail-3}
\end{figure}

\begin{figure}[!t]
 \centering
 \subfigure[System Track Non-participants]{
  \includegraphics[width=0.77\textwidth,viewport=10 0 330 300]{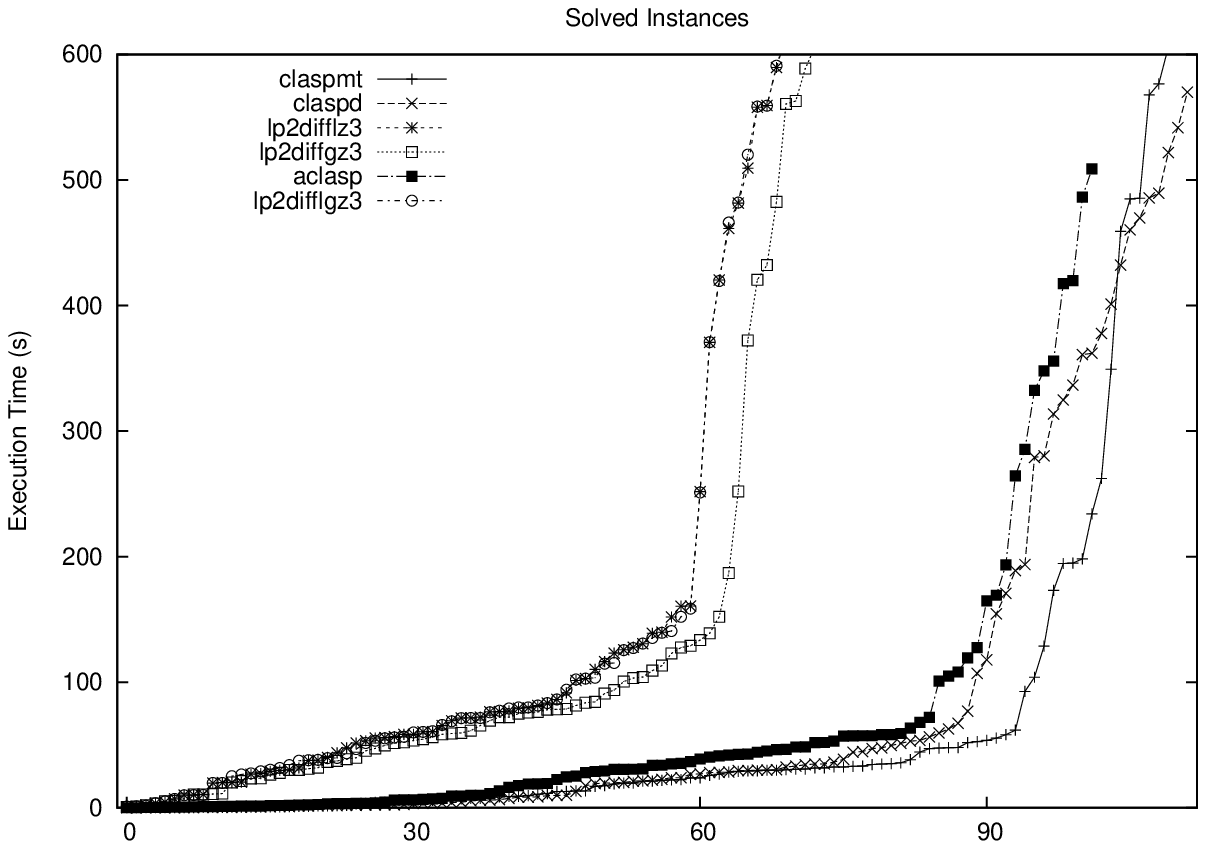}\label{fig:plots-detail-nonp}}
 \subfigure[Team Track Optimization]{
  \includegraphics[width=0.77\textwidth,viewport=10 0 330 300]{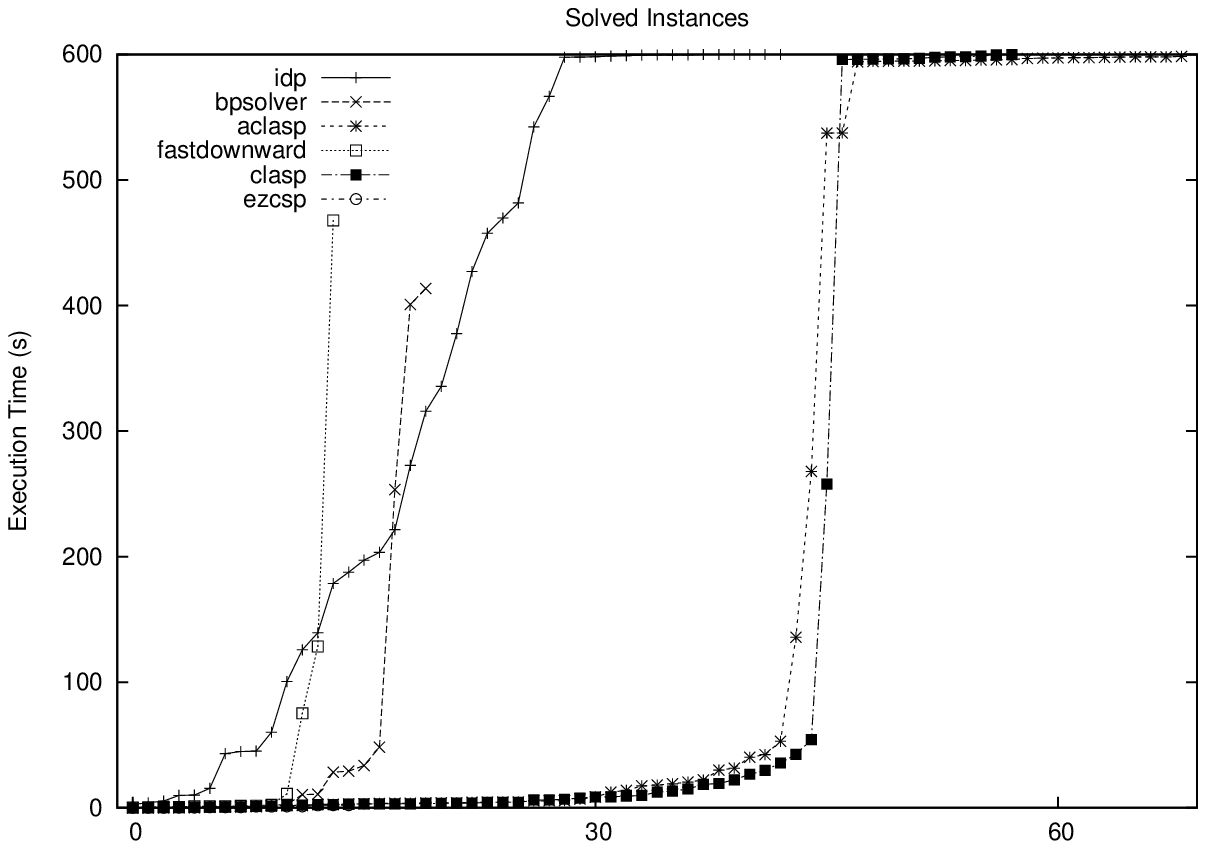}\label{fig:plots-detail-opt}}\\
 \vspace*{1ex}
 \caption{Results in Detail: Execution time (y-axis), Solved Instances (x-axis). }\label{fig:plots-detail-4}
\end{figure}

\nopagebreak[4]
\end{document}